\documentclass[10pt,twocolumn,letterpaper]{article}
\usepackage{iccv}
\usepackage{times}
\usepackage{epsfig}
\usepackage{graphicx}
\usepackage{amsmath}
\usepackage{amssymb}
\usepackage{color}
\usepackage[dvipsnames]{xcolor}
\usepackage[T1]{fontenc}
\usepackage{array}
\usepackage{svg}
\usepackage{placeins}
\usepackage{import}
\usepackage{xifthen}
\usepackage{pdfpages}
\usepackage{transparent}
\usepackage{pifont}
\usepackage{xspace}
\usepackage{multirow}
\usepackage{etoolbox,xpatch}
\usepackage{rotating}
\usepackage{enumitem}
\usepackage{pdflscape}
\usepackage{algorithm}
\usepackage{algpseudocodex}
\usepackage[all=normal, mathspacing=tight,mathdisplays=normal, floats=tight, paragraphs=tight, lists=normal]{savetrees}

\usepackage{tabularx, colortbl, booktabs}

\usepackage[numbers,sort,compress]{natbib}

\usepackage[pagebackref=true,breaklinks=true,letterpaper=true,colorlinks,bookmarks=false]{hyperref}

\AtBeginEnvironment{minted}{\dontdofcolorbox}
\def\dontdofcolorbox{\renewcommand\fcolorbox[4][]{##4}}
\xpatchcmd{\inputminted}{\minted@fvset}{\minted@fvset\dontdofcolorbox}{}{}
\renewcommand{\b}{\boldsymbol}
\newcommand{\m}{\mathcal}

\DeclareMathOperator*{\argmax}{arg\,max}

\definecolor{ood}{rgb}{0.96,1,0.94}
\definecolor{idmiscls}{rgb}{1,0.97,0.96}
\definecolor{light-gray}{gray}{0.7}
\definecolor{dark-gray}{gray}{0.4}
\newcolumntype{a}{>{\columncolor{ood}}l}
\newcolumntype{b}{>{\columncolor{idmiscls}}l}

\usepackage[capitalize]{cleveref}
\crefname{section}{Sec.}{Secs.}
\Crefname{section}{Section}{Sections}
\Crefname{table}{Table}{Tables}
\crefname{table}{Tab.}{Tabs.}
\crefname{algorithm}{Alg.}{Algs.}

\def\Highlight{\leavevmode\rlap{\hbox to \hsize{\color{lightgray!50}\leaders\hrule height .8\baselineskip depth .5ex\hfill}}}

\iccvfinalcopy 

\begin{document}

\title{Window-Based Early-Exit Cascades for Uncertainty Estimation: \\When Deep Ensembles are More Efficient than Single Models}

\author{Guoxuan Xia  \\
Imperial College London \\
{\tt\small g.xia21@imperial.ac.uk}
\and
Christos-Savvas Bouganis \\
Imperial College London\\
{\tt\small christos-savvas.bouganis@imperial.ac.uk}
}

\maketitle
\ificcvfinal\thispagestyle{empty}\fi

\begin{abstract}
Deep Ensembles are a simple, reliable, and effective method of improving both the predictive performance and uncertainty estimates of deep learning approaches. However, they are widely criticised as being computationally expensive, due to the need to deploy multiple independent models. 
Recent work has challenged this view, showing that for predictive accuracy, ensembles can be more computationally efficient (at inference) than scaling single models within an architecture family. This is achieved by \emph{cascading} ensemble members via an \emph{early-exit} approach. 
In this work, we investigate extending these efficiency gains to tasks related to uncertainty estimation. As many such tasks, \eg selective classification, are binary classification, our key novel insight is to only pass samples within a \emph{window} close to the binary decision boundary to later cascade stages.   
Experiments on ImageNet-scale data across a number of network architectures and uncertainty tasks show that the proposed window-based early-exit approach is able to achieve a superior uncertainty-computation trade-off compared to scaling single models. For example, a cascaded EfficientNet-B2 ensemble is able to achieve similar coverage at 5\% risk as a single EfficientNet-B4 with <30\% the number of MACs. We also find that cascades/ensembles give more \emph{reliable} improvements on OOD data vs scaling models up. Code for this work is available at:\\ \url{https://github.com/Guoxoug/window-early-exit}.
\end{abstract}

\section{Introduction}
\begin{figure}
    \centering
    \includegraphics[width=\linewidth]{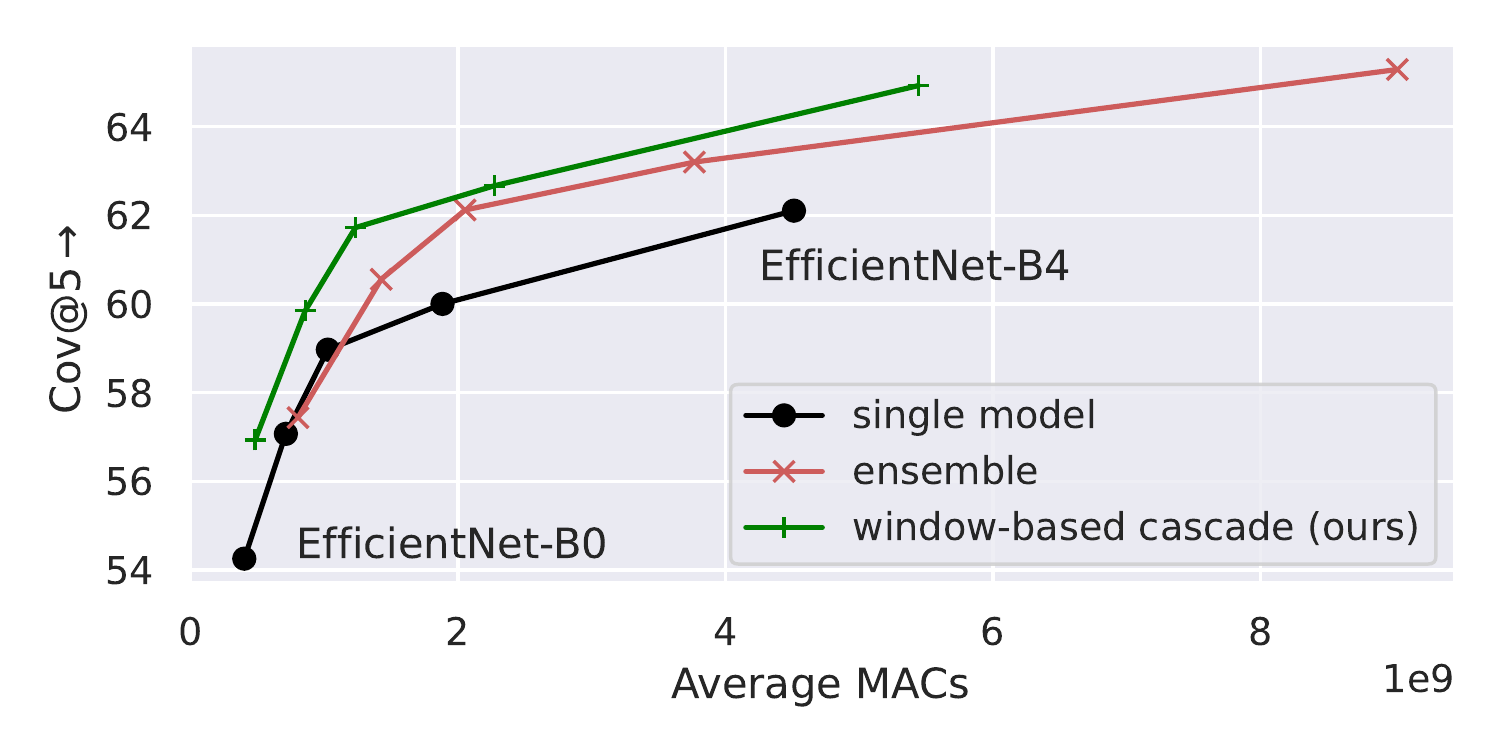}
    \caption{Coverage at risk=5\% (selective classification performance) against average computation per sample (MACs). Ensembles are able to outperform scaling single models at higher levels of computation. \textbf{Our \textcolor{ForestGreen}{window-based cascade} approach is able to achieve a better uncertainty-computation trade-off compared to single models} via adaptive inference. The models are EfficientNets, ensembles are size $M=2$ and the dataset is ImageNet-1k.}
    \label{fig:tease}
\end{figure}
It is important for deep learning systems to provide reliable estimates of predictive uncertainty in scenarios where it is costly to make mistakes, \eg healthcare \cite{medunc} and autonomous driving \cite{avunc}. A model should know when to defer a prediction to a human expert or simply discard it as too uncertain. Deep Ensembles \cite{Lakshminarayanan2017SimpleAS}, where multiple networks of the same architecture are trained independently and combined at inference, are a simple approach to improving both the predictive and uncertainty-related performance of deep neural networks. Moreover, as it is an easy-to-implement baseline, the approach has accumulated a large volume of empirical evidence in support of its efficacy across a wide range of data types and uncertainty-related tasks \cite{ovadia2019can, Malinin2020EnsembleDD, Xia2022OnTU, Malinin2021UncertaintyEI, gustafsson2020evaluating, LDU, Kim2021AUB, Fort2019DeepEA, Beluch2018ThePO, malinin2021shifts, Wu2020EnsembleAF, Fathullah2020EnsembleDA, geifman2018biasreduced,logit-based-edd}. 

A common criticism of Deep Ensembles is that they are expensive to deploy \cite{Malinin2020EnsembleDD, havasi2021training, Fathullah2020EnsembleDA}, as the cost scales linearly with the ensemble size. However, recent work \cite{powerlaw, Kondratyuk2020WhenES, Zhao2020TowardBA, pmlr-v157-deng21a, wang2022wisdom}, has shown that \textit{ensembling can be more efficient than scaling a single model} when evaluating for predictive accuracy. In particular, \citet{wang2022wisdom} show that cascading ensemble members via adaptive early-exiting \cite{Teerapittayanon2016BranchyNetFI,Wang2017IDKCF,adaptive} enables ensembles to obtain consistently better accuracy-computation trade-offs vs scaling single models within an architecture family. This approach involves sequentially running inference on each member and exiting computation early if the ensemble \textit{up to that point} is sufficiently confident. Thus computation is saved by only passing the most difficult samples to the more powerful later cascade stages. Although \citet{wang2022wisdom} have shown promising results for accuracy, the uncertainty-computation trade-off for ensembles remains unexplored. The \textbf{key contributions} of this work are:
\begin{enumerate}
    \item We investigate the unexplored trade-off between \textit{uncertainty-related} performance and computational cost for Deep Ensembles vs scaling single models. We evaluate over multiple downstream tasks: selective classification (SC) \cite{Geifman2017SelectiveCF}, out-of-distribution (OOD) detection \cite{yang2021oodsurvey}, and selective classification with OOD data (SCOD) \cite{Kim2021AUB, Xia_2022_ACCV}. Experiments are performed over a range of ImageNet-scale datasets and convolutional neural network (CNN) architectures. Compared to single-model scaling, we find that Deep Ensembles are more efficient for high levels of compute and give more \textit{reliable} improvements on OOD data.
    
    \item We propose a novel \textit{window-based} early-exit policy, which enables cascaded Deep Ensembles to achieve a \textit{superior uncertainty-computation trade-off} compared to model scaling. Our key insight is that many downstream tasks for uncertainty estimation can be formulated as binary classification. Thus, only samples in a \textit{window} near the binary decision boundary should be evaluated by the more costly but better-performing later cascade stages. Results can be previewed in \cref{fig:tease}. We also validate our exit policy on MSDNet \cite{Huang2018MultiScaleDN}, a dedicated multi-exit architecture.
\end{enumerate}
\section{Preliminaries}\label{prelims}
\subsection{Uncertainty estimation: downstream tasks}\label{sec:tasks}
\begin{figure}
    \centering
    \includegraphics[width=\linewidth]{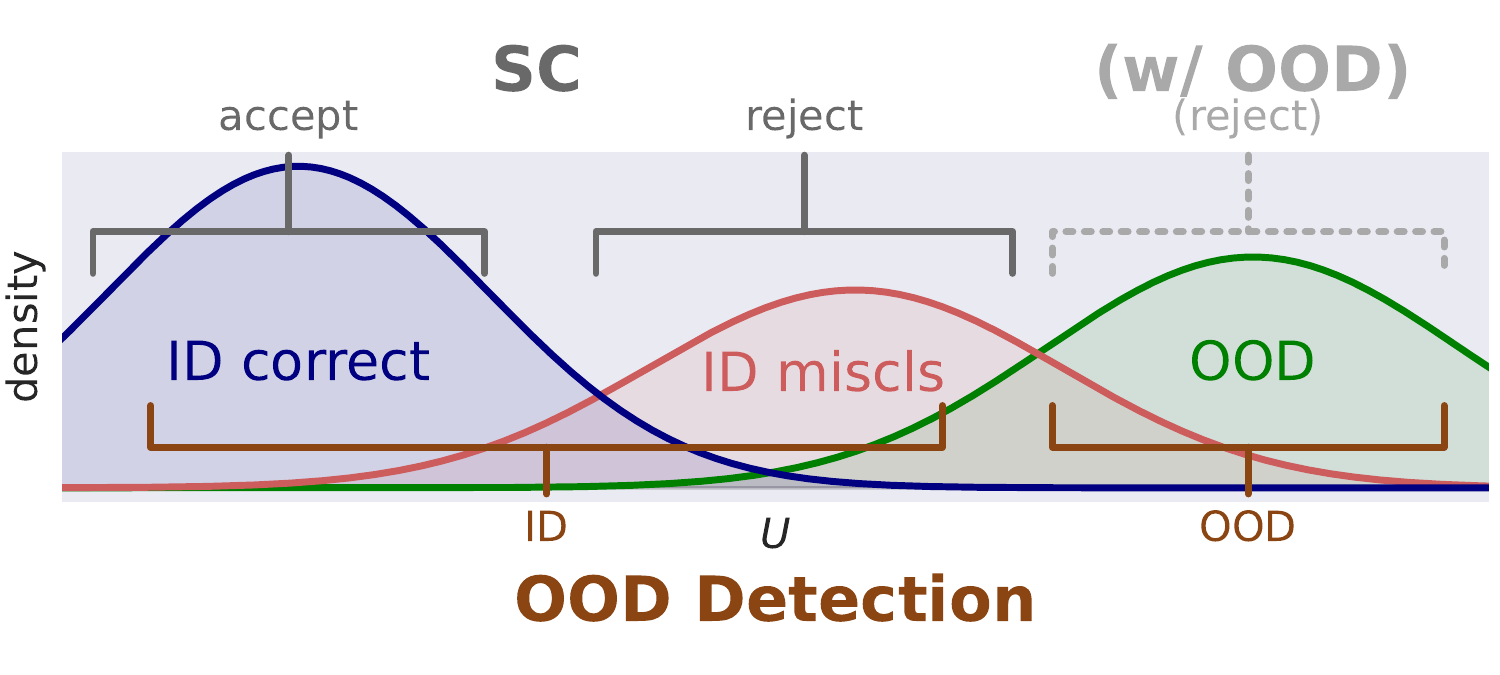}
    \caption{Illustration of uncertainty-related tasks evaluated in this work. \textbf{\textcolor{dark-gray}{SC}} aims to accept correct predictions and reject misclassifications. \textbf{\textcolor{Brown}{OOD detection}} aims to separate ID data from OOD data. \textbf{\textcolor{dark-gray}{SCOD}} is \textbf{\textcolor{dark-gray}{SC}} where we additionally want to reject OOD data.}
    \label{fig:illustps}
\end{figure}

In this work, we take a task-oriented view, that motivates the estimation of uncertainties from the perspective of wanting to detect costly or failed predictions \cite{Xia_2022_ACCV, jaeger2023a}. \cref{fig:illustps} illustrates the tasks discussed in this section.
\paragraph{Selective classification (SC).} 
Consider a neural network classifier, with parameters $\b \theta$, that models the conditional distribution $P(y|\b x;\b \theta)$ over $K$ labels $y \in \m Y = \{\omega_k\}_{k=1}^K$ given inputs $\b x \in \m X = \mathbb R^D$. It is trained on dataset $\m D_\text{tr} = \{y^{(n)},\b x^{(n)}\}_{n=1}^{N}$ drawn from the distribution $p_\text{ID}(y,\b x)$. This is typically achieved with a categorical softmax output. The classification function $f$ for test input $\b x^*$ is defined as
\begin{equation}\label{eq:classifier}
     f(\b x^*) =\hat y  = \argmax_\omega P(\omega|\b x^*;\b \theta)~.
\end{equation}
A \textit{selective classifier} \cite{ElYaniv2010OnTF} is a function pair $(f, g)$  that includes the aforementioned classifier $f(\b x)$, which yields prediction $\hat y$, and a \textit{binary} rejection function, 
\begin{equation}\label{eq:rej}
    g(\b x;\tau) = \begin{cases}
    0\text{ (reject prediction)}, &\text{if }U(\b x) > \tau\\
    1\text{ (accept prediction)}, &\text{if }U(\b x) \leq \tau~,
    \end{cases}
\end{equation}
where $U(\b x)$ is a measure of predictive uncertainty and $\tau$ is an operating threshold. $U$ is typically calculated using the outputs of the neural network. Intuitively, uncertain samples are rejected. The selective risk \cite{Geifman2017SelectiveCF} is then defined as
\begin{equation}\label{risk}
    \text{Risk}(f,g;\tau) = \frac{\mathbb E_{p_\text{ID}(\b x)}[g(\b x;\tau)\m L_\text{SC}(f(\b x))]}{\mathbb E_{p_\text{ID}(\b x)}[g(\b x;\tau)]}~,
\end{equation}
where 
\begin{equation}\label{eq:loss}
    \m L_\text{SC}(f(\b x)) = \begin{cases}
    0, &\text{if } f(\b x) = y \quad \text{(correct)}\\
    1, &\text{if } f(\b x) \neq y\quad \text{(misclassified)}~,
    \end{cases}
\end{equation}
is the 0/1 error. Intuitively, selective risk measures the average loss over the \textit{accepted} samples. The coverage is defined as $\text{Cov} = \mathbb E_{p_\text{ID}(\b x)}[g(\b x;\tau)]$, i.e. the proportion of predictions accepted. It is desirable to maximise coverage (accept as many as possible) and minimise risk (reduce the error rate of accepted samples). This is achieved by either improving $g$, i.e. better rejection of misclassifications, or by improving $f$, i.e. better accuracy so fewer misclassifications to reject. 

SC performance can be either measured using the threshold-free metric, Area Under the Risk-Coverage curve (AURC$\downarrow$)\footnote{Arrows by a metric indicate whether higher ($\uparrow$) or lower ($\downarrow$) is better.} \cite{Geifman2017SelectiveCF}, or be evaluated at an appropriate threshold $\tau$ based on a desired criterion \cite{SC_VQA}, \eg coverage at 5\% risk (Cov@5$\uparrow$). In practice, $\tau$ can be determined by matching the desired criterion on a held-out validation dataset.
\paragraph{Out-of-distribution (OOD) detection.}
Regular classification assumes that at deployment, samples originate from the same distribution as the one the classifier was trained on,  $(y^*, \b x^*) \sim p_\text{ID}(y,\b x)$. This may not be the case in the real world. OOD detection aims to filter out samples that originate from some distribution $p_\text{OOD}(\b x)$ \textit{with labels that are disjoint from} $\mathcal Y$ \cite{yang2021oodsurvey}.\footnote{Note this is different to \textit{covariate shift} where $\m Y$ is preserved \cite{yang2021oodsurvey}} We now assume a mix of in-distribution (ID) and OOD data occurs at deployment,
\begin{align}\label{eq:mix}
    \b x^* &\sim p_\text{mix}(\b x) \nonumber \\p_\text{mix}(\b x) &= \alpha p_\text{ID}(\b x) + (1-\alpha)p_\text{OOD}(\b x), \quad \alpha \in [0,1]~.
\end{align}
Predictions on OOD data may be costly to pass downstream, \eg a self-driving car classifying a bear as a car. OOD detection is again performed using a \textit{binary} detection function,
\begin{equation}\label{eq:bin_OOD}
    g(\b x;\tau) = \begin{cases}
    0\text{ (OOD)}, &\text{if }U(\b x) > \tau\\
    1\text{ (ID)}, &\text{if }U(\b x) \leq \tau~.
    \end{cases}
\end{equation}
We intuitively want high uncertainty on OOD data. OOD detection may be evaluated using the threshold-free Area Under the Receiver Operating Characteristic curve (AUROC$\uparrow$), or at a threshold $\tau$ determined on ID data \eg false positive rate at 95\% true positive rate (FPR@95$\downarrow$). We assume no access to OOD data before deployment, although other work \cite{hendrycks2019oe, sehwag2021ssd} relaxes this constraint.

\paragraph{Selective classification with OOD Data (SCOD).} As OOD detection does not consider the quality of predictions classified as ``ID'' by the detector (there may be many ID misclassifications), recent work \cite{Kim2021AUB, Xia_2022_ACCV, jaeger2023a} has argued for a task that combines SC with OOD detection. We adopt the problem definition and naming from \cite{Xia_2022_ACCV}.\footnote{The research community has yet to converge on a single name for this task, so other than ``SCOD'' readers may encounter ``failure detection'' \cite{jaeger2023a}, ``unknown detection'' \cite{Kim2021AUB} or ``unified open-set recognition'' \cite{cen2023the}.} In SCOD, the objective is to reject \textit{both} ID misclassifications \textit{and} OOD samples, in order to minimise the loss on the accepted predictions.  We assume \cref{eq:mix} at deployment and perform binary rejection using \cref{eq:rej}. The loss becomes 
\begin{equation}\label{eq:scodloss}
    \m L_\text{SCOD}(f(\b x)) = \begin{cases}
    0, &\text{if } f(\b x) = y,  (y, \b x) \sim p_\text{ID} \\
    1, &\text{if } f(\b x) \neq y,  (y, \b x) \sim p_\text{ID} \\
    \beta, &\text{if } \b x \sim p_\text{OOD}, \quad (\beta \geq 0)~,
    \end{cases}
\end{equation}
where a cost is incurred if either an ID misclassification or an OOD sample is accepted, and $\beta$ reflects the relative cost of these two events. The risk is now given by 
\begin{equation}\label{eq:scodrisk}
    R(f,g;\tau) = \frac{\mathbb E_{p_\text{mix}(\b x)}[g(\b x;\tau)\m L_\text{SCOD}(f(\b x))]}{\mathbb E_{p_\text{mix}(\b x)}[g(\b x;\tau)]}~.
\end{equation}
SCOD can be evaluated similarly to SC with AURC$\downarrow$. However, as we assume no access to $p_\text{OOD}$ pre-deployment, we cannot set threshold $\tau$ according to risk on $p_\text{mix}$ (\cref{eq:scodrisk}). We can however set $\tau$ according to coverage and then use \eg Risk@(Cov=)80$\downarrow$ as a threshold-specific measure. This coverage may be only over the ID data (equivalent to TPR for OOD detection), in which case an ID validation set can be used. Alternatively, one may use coverage over all samples. Here statistics of $U$ (not labels) would need to be obtained from $p_\text{mix}$ during deployment.
\subsection{Ensembles and cascades}
\paragraph{Deep Ensembles.}
Training multiple deep neural networks of the same architecture on the same data, but with different random seeds is a well-established approach to improving the quality of uncertainty estimates \cite{Lakshminarayanan2017SimpleAS}. With a Deep Ensemble of $M$ networks $\Theta = \{\b \theta^{(m)}\}_{m=1}^M$, the predictive distribution is typically obtained as
\begin{equation}\label{eq:pred}
    P(y|\b x; \Theta) = \frac{1}{M}\sum_{m=1}^M P(y|\b x;\b \theta^{(m)})~,
\end{equation}
that is to say the mean of the member softmax outputs. Predictions are then obtained similarly to \cref{eq:classifier}. To calculate ensemble uncertainty we adopt the simple approaches of directly using the predictive distribution or averaging individual member uncertainties as in \cite{Xia2022OnTU},
\begin{align}
    U(\b x; \Theta) &= U\left(P(y|\b x; \Theta)\right)~,\label{eq:ens_unc1} \\
    U(\b x; \Theta) &= \frac{1}{M}\sum_{m=1}^M U(\b x;\b \theta^{(m)})~,\label{eq:ens_unc2}
\end{align}
although there exist more advanced methods \cite{Depeweg2018DecompositionOU, Malinin2021UncertaintyEI, Xia2022OnTU}.

\paragraph{Adaptive inference with early-exit cascades.} 
One approach to improving the efficiency of deep learning approaches is through \textit{adaptive inference}, where computation is allocated conditionally based on the input sample \cite{adaptive, dynamicsurvey}. Early-exit approaches \cite{adaptive} involve terminating computation early depending on the input when producing a prediction. This may be achieved by designing architectures with intermediate predictors \cite{Teerapittayanon2016BranchyNetFI, Huang2018MultiScaleDN}, or by cascading full networks in series \cite{Wang2017IDKCF, wang2022wisdom}. By placing more powerful, but also more expensive predictors later on, computation is intuitively saved on ``easy'' samples that exit early. 
\begin{algorithm}[b]
\caption{Cascaded Deep Ensemble Classifier}\label{alg:cascade}
\small
\begin{algorithmic}
\Require Ensemble $\{\b \theta^{(m)}\}_{m=1}^M$ \quad Thresholds $\{t^{(m)}\}_{m=1}^{M-1}$ \\Test Input $\b x^*$
\For{$l=1,2,\dots,M$}
\State Do inference on $\b x^*$ using $l$th ensemble member $\b \theta^{(l)}$
\State Cache outputs for $l$th member
\State Calculate $U(\b x^*; \{\b \theta^{(m)}\}_{m=1}^l)$ 
\LComment{\cref{eq:ens_unc1,eq:ens_unc2} using model outputs only up to $l$}
\If{$l=M$ or $U < t^{(l)}$} 
\Comment{single threshold (per exit)}
    \LComment{exit early using $l$ members if confident, or full ens.}
    \State \Return $\hat y = \argmax_\omega P(\omega|\b x^*; \{\b \theta^{(m)}\}_{m=1}^l)$
\EndIf
\EndFor
\end{algorithmic}
\end{algorithm}
\begin{figure*}[h]
    \centering
    \includegraphics[width=\textwidth]{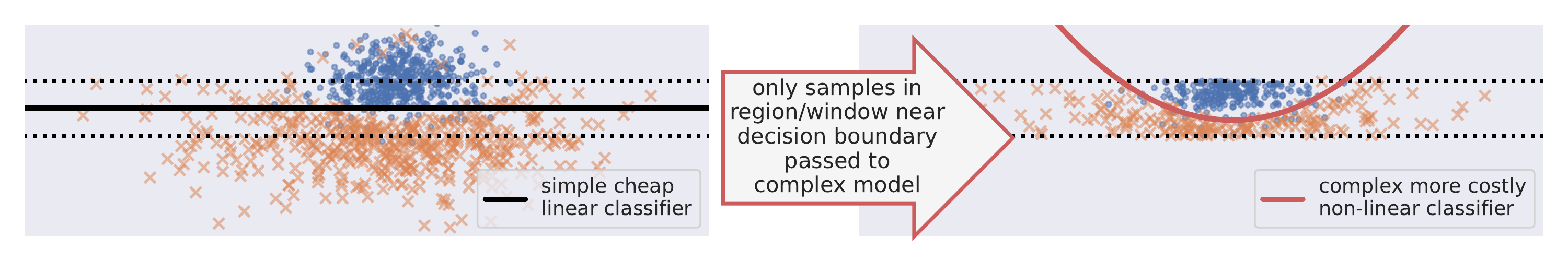}
    \caption{Illustration of intuition for early-exit approaches with a 2D binary classification problem. \textbf{The most efficient gains in accuracy are obtained when samples in a region/window near the decision boundary of the first, simple classifier are passed to the second, more powerful (and more costly) classifier.} These samples are most likely to be 1) incorrectly classified initially and 2) corrected later on.}
    \label{fig:ee_int}
\end{figure*}
Although there is work on learnt exit policies \cite{chen2020learning, learnexit}, a simple and widely adopted approach is to exit if intermediate estimates of uncertainty are below a threshold $t$ \cite{Teerapittayanon2016BranchyNetFI, Wang2017IDKCF, Kouris2021MultiExitSS, Huang2018MultiScaleDN}. If there are multiple exits, then there may be multiple thresholds. Thresholds can be chosen arbitrarily or optimised on an ID validation set subject to an accuracy or computational constraint \cite{Tan_Li_Wang_Huang_Xu_2021, wang2022wisdom, Kouris2021MultiExitSS}. \cref{alg:cascade} applies this to Deep Ensembles (for classification) \cite{wang2022wisdom}.

In their work, \citet{wang2022wisdom} show that the approach in \cref{alg:cascade} enables ensembles to achieve a better accuracy-computation trade-off compared to scaling single models. We will now extend this to uncertainty estimation.

\begin{figure}[]
    \centering
    \includegraphics[width=\linewidth]{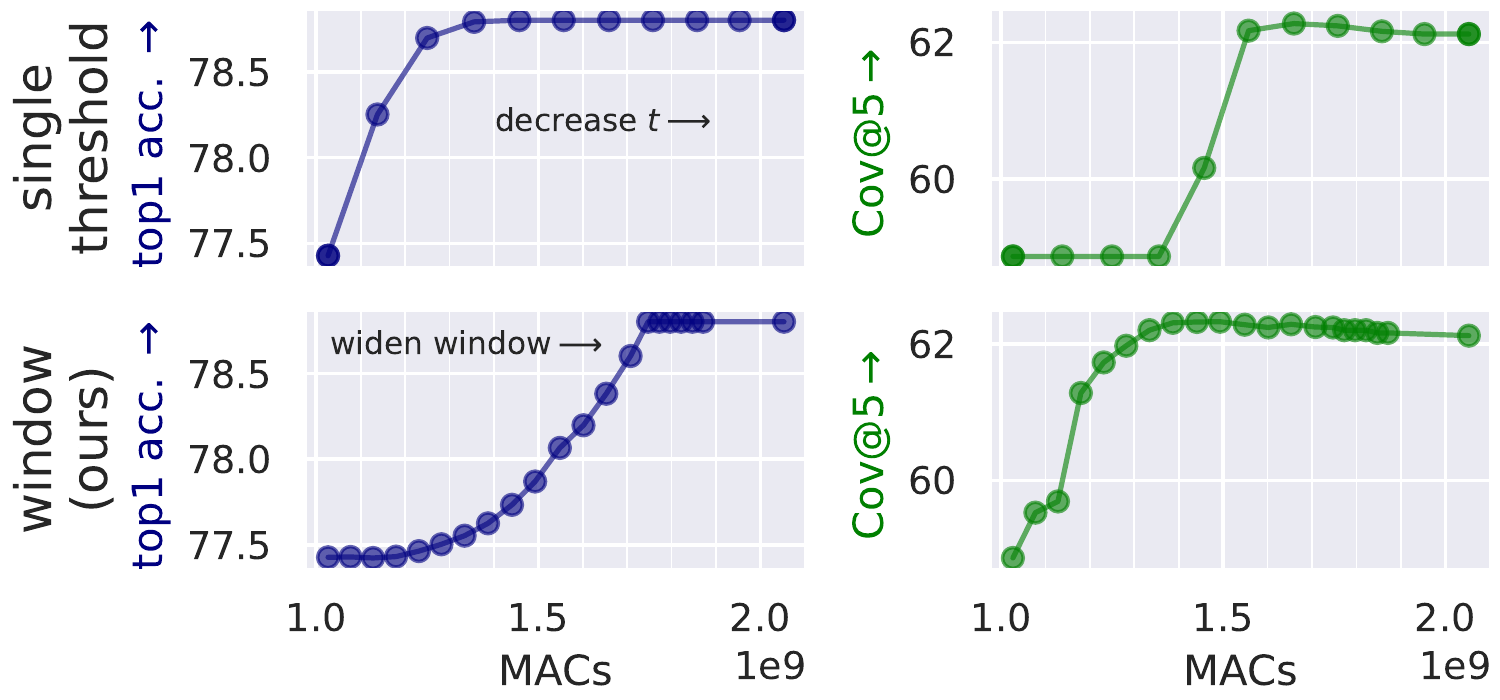}
    \caption{The existing single-threshold policy (top) from \cref{alg:cascade} vs our window-based approach (bottom) (\cref{alg:unc_cascade}), as more samples are passed onto the second stage of an ensemble, increasing computation. \textbf{Although a single threshold is efficient for accuracy, it is not for selective classification (\textcolor{ForestGreen}{Cov@5$\uparrow$})}. \textbf{Our window-based policy shows efficient improvements on the uncertainty-related task}, by targeting the accept/reject decision boundary. For SC we use MSP ($U(\b x)= -\max_\omega P(\omega|\b x)$) and the final exited uncertainties (\cref{fig:framework_illust}) with \cref{eq:ens_unc1}. We use an EfficientNet-B2 ensemble with $M=2$. The data is ImageNet-1k.}
    \label{fig:strats}
\end{figure}

\section{Window-Based Cascades: \\Targeting Uncertainty-Related Tasks}
\cref{alg:cascade} (single threshold) results in underwhelming uncertainty efficiency. \cref{fig:strats} shows that although a rapid improvement in predictive accuracy can be achieved with a small increase in average multiply–accumulate (MAC) operations, uncertainty-related Cov@5$\uparrow$ only improves after a much larger increase in computation. To better understand why we consider one intuition for early exiting.

\subsection{Early-exiting on samples far away\\from the decision boundary}
\begin{figure*}[t]
\footnotesize
\textsf{
    \centering
    \def\svgwidth{\linewidth}
    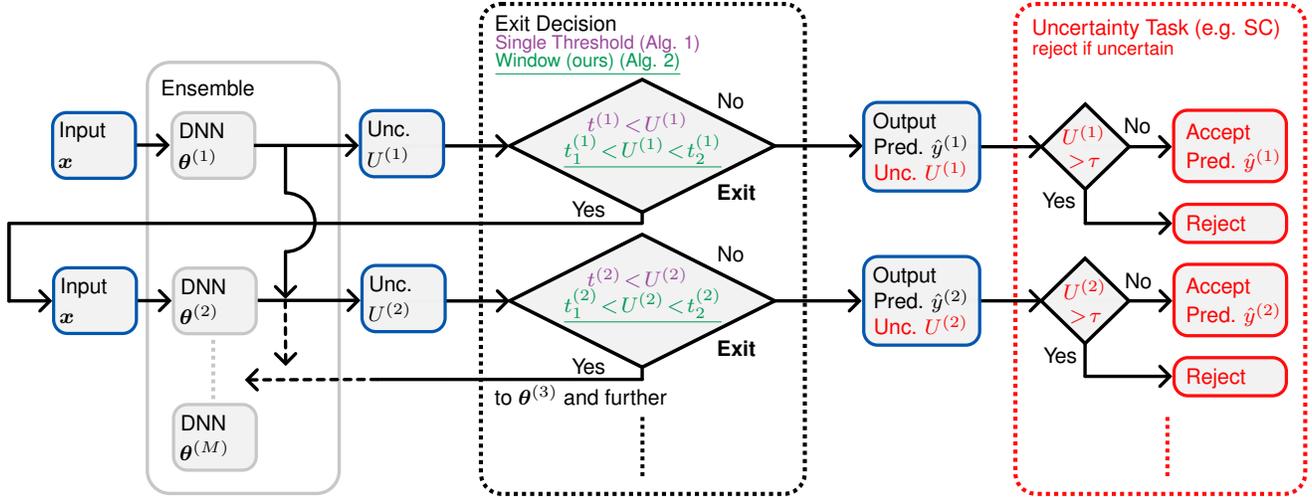

}
    \caption{End-to-end illustration of cascaded Deep Ensembles applied to a binary uncertainty-related task. Each ensemble member is run sequentially. Computation terminates depending on a \textcolor{Purple}{single threshold} (\cref{alg:cascade}) or \textcolor{ForestGreen}{\textbf{\underline{window-based (ours)}}} (\cref{alg:unc_cascade}) exit decision. After exiting, the downstream \textcolor{red}{uncertainty-related task} is performed using the uncertainty estimate of the ensemble \textit{up to the exit point}.}
    \label{fig:framework_illust}
\end{figure*}
The typical intuition behind explaining early-exit approaches for multi-class classification is that we can save computation by exiting if an intermediate classifier is confident (low uncertainty) in its prediction. However, it is non-obvious as to how to extend this intuition to uncertainty estimation. Do we need a ``meta-uncertainty'' to measure the uncertainty of uncertainty estimates? Our key insight is to view early-exiting from a \textit{decision boundary} perspective, and to recall that many downstream tasks using estimates of predictive uncertainty are in fact \textit{binary classification} (see \cref{sec:tasks}, \cref{eq:bin_OOD,eq:rej}).

Common measures of uncertainty used for early-exiting, such as the maximum softmax probability (MSP) \cite{hendrycks17baseline}, intuitively approximate the distance from the classifier decision boundary \cite{Pearce2021UnderstandingSC}. Passing samples from the region \textit{near the decision boundary} to a more powerful model will result in the most efficient improvement in classification performance. This is because these samples are more likely to be:
\begin{enumerate}
    \item incorrectly classified by the initial stage and 
    \item corrected by the next stage (given they are wrong).
\end{enumerate}
This is illustrated in \cref{fig:ee_int}. Applying this intuition to uncertainty-related tasks leads us to our proposed approach.
\subsection{Approach: window-based early-exiting}\label{sec:approach}
In contrast to existing early-exit approaches that target the decision boundary of the multi-class classification problem using a single threshold on uncertainty,
\textbf{we propose to early-exit only on samples outside of a \textit{window} $[t_1, t_2]$ around the decision boundary $\tau$ }(\cref{eq:bin_OOD,eq:rej}) corresponding to the uncertainty-related task of interest, as in \cref{fig:ee_int}. Our approach is detailed in \cref{alg:unc_cascade} and \cref{fig:framework_illust}. 
\begin{algorithm}
\caption{Window-Based Cascaded Deep Ens. (ours)}\label{alg:unc_cascade}
\small
\begin{algorithmic}
\Require Ensemble $\{\b \theta^{(m)}\}_{m=1}^M$ \quad Windows $\{[t_1,t_2]^{(m)}\}_{m=1}^{M-1}$\\Test Input $\b x^*$
\For{$l=1,2,\dots,M$}
\State Do inference on $\b x^*$ using $l$th ensemble member $\b \theta^{(l)}$
\State Cache outputs for $l$th member
\State Calculate $U(\b x^*; \{\b \theta^{(m)}\}_{m=1}^l)$ 
\LComment{\cref{eq:ens_unc1,eq:ens_unc2} using model outputs only up to $l$}
\BeginBox[fill=lightgray!20]
\If{$l=M$ or $U \notin [t_1,t_2]^{(l)}$} 
\LComment{exit if $U$ outside window, or full ens.}
\EndBox
\State \Return $U$, $\hat y = \argmax_\omega P(\omega|\b x^*; \{\b \theta^{(m)}\}_{m=1}^l)$\EndIf
\EndFor
\end{algorithmic}
\end{algorithm}
\paragraph{Implementation details.}
 In our case, to find $[t_1, t_2]$, we find $\tau$ for each exit independently on an ID validation dataset according to the desired criterion, e.g. 80\% coverage (\cref{sec:tasks}), and draw a window to capture a certain symmetric percentile of samples either side of $\tau$. For example, setting $[t_1,t_2]$ at $\pm15$ percentiles around $\tau$ would result in 70\% of samples exiting early. Final binary classification for the uncertainty-related task is then performed using a new $\tau$ which is found using the final exited uncertainties of the overall adaptive system (\cref{fig:framework_illust}).

We note that it is possible to optimise the window parameters $[t_1, t_2]$ on the ID validation dataset for further improvements, just as regular early-exit thresholds can be optimised \cite{Tan_Li_Wang_Huang_Xu_2021, wang2022wisdom, Kouris2021MultiExitSS}. This becomes more relevant in systems with multiple exits. It is desirable to automatically determine $[t_1, t_2]$ when the number of thresholds to find is higher. However, we leave this to future work, as our aim is to simply show the efficacy of our window-based approach.
\begin{figure}
    \centering
    \includegraphics[width=\linewidth]{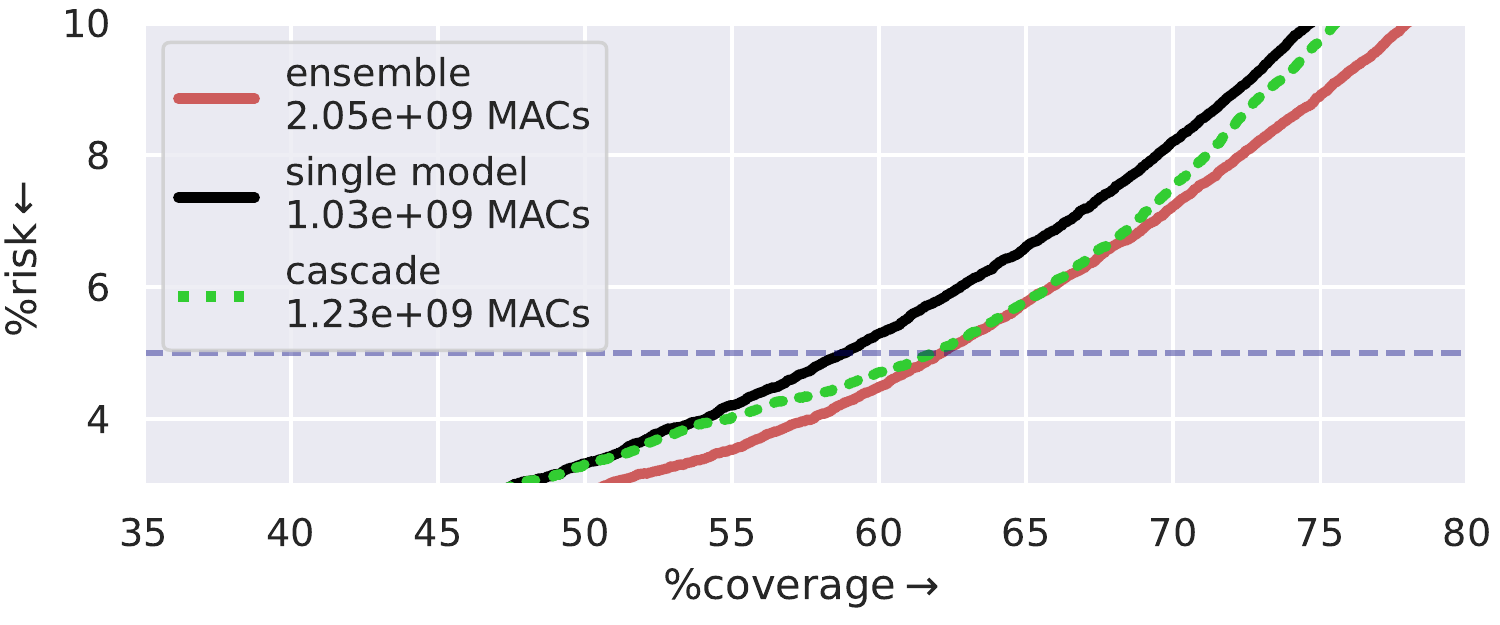}
    \caption{Risk-coverage curves for single model, ensemble, and a window-based cascade. The cascade aims to target risk=5\%. \textbf{Near the desired operating point the cascade is able to achieve ensemble-like coverage performance}, for a small increase in inference MACs. Further away, at \textit{irrelevant} operating points, the cascade behaves closer to the single model. This is because the uncertainties around these operating points will be mainly from just the single model, not the ensemble. The setup is the same as \cref{fig:strats}.}
    \label{fig:touching}
\end{figure}
\paragraph{Efficient gains in uncertainty estimation.}
\cref{fig:strats} shows that this window-based approach efficiently improves Cov@5$\uparrow$. We note that gains in accuracy (over all samples) are worse, as the multi-class decision boundary is no longer targeted. However, during SC, we are concerned with the accuracy of \textit{only the accepted samples}. To explain the SC behaviour of single threshold (\cref{alg:cascade}) in \cref{fig:strats}, imagine sweeping a single dashed line over \cref{fig:ee_int} -- the step-like boost occurs when the single $t$ passes over the decision boundary $\tau$ meaning samples near $\tau$ are now passed on.

The intuition so far has only been linked to the idea that later stages of the cascade have a better binary classifier $g$, however for SC, it is also the case that they may have a better multi-class classifier $f$. In this case, after progressing to the next stage, samples may find themselves changing from misclassifications to correct (multi-class) predictions.

Our approach targets a \textit{specific threshold}, which is the \textit{actual deployment setting}. The efficiency gains arise from the fact that \textit{only} samples within a window around $\tau$ are passed to the next cascade stage (\cref{fig:ee_int}) -- the most relevant samples receive additional computation. \cref{fig:touching} shows that a window-based cascade is able to achieve the full ensemble's performance around the desired 5\% risk, however, its behaviour tends to the single model further away (outside the window). This is because, for decision boundaries outside the window, the sample uncertainties will be from the earlier stage. As such, threshold-free evaluation metrics (\eg AURC$\downarrow$) are not suitable for a given window.

\section{Experimental Results and Discussion}
\subsection{Main results}

\paragraph{Setup.}
We train two families of computationally efficient image classification CNNs on ImageNet-1k \cite{Russakovsky2015ImageNetLS}: EfficientNet \cite{pmlr-v97-tan19a} and MobileNet-V2 \cite{Sandler2018MobileNetV2IR}. For EfficientNet, we scale from B0$\rightarrow$B4. For MobileNet-V2 we use the scalings\footnote{\url{https://github.com/keras-team/keras/blob/master/keras/applications/mobilenet_v2.py}.} in \cref{tab:mob_scaling}. For each model we train a Deep Ensemble size $M=2$ using random seeds $\{1,2\}$. This results in 10 individual ensembles, 5 composed of EfficientNets and 5 composed of MobileNet-V2s. We train for 250 epochs, holding out a random subset of 50,000 images for validation, and then evaluate on the original validation set.
\begin{table}[h]
\caption{MobileNet-V2 scaling used in experiments.}
\label{tab:mob_scaling}
\small
\centering
\begin{tabular}{@{}llllll@{}}
\toprule
input resolution & 160 & 192 & 224 & 224 & 224 \\
width factor     & 1.0 & 1.0 & 1.0 & 1.3 & 1.4 \\ \bottomrule
\end{tabular}
\end{table}

All models are trained using Pytorch \cite{torch} and Lightning \cite{Falcon_PyTorch_Lightning_2019}. Full training details and code can be found in \cref{app:setup}. We measure inference computation in multiply-accumulate (MAC) operations\footnote{Other work may refer to MACs as FLOPs or MADs.} using ptflops.\footnote{\url{https://github.com/sovrasov/flops-counter.pytorch}} 

We cascade each ensemble in both directions over seeds ($1\rightarrow\{1,2\}$ and $2\rightarrow\{1,2\}$) and report the mean values. Note that from this point onwards we refer to cascaded Deep Ensembles as ``cascades'' and non-adaptive Deep Ensembles as simply ``ensembles'' for the sake of brevity.
\paragraph{Scaling vs architecture design.} We note that the main comparison in this section is between ensembling and \textit{simple scaling}, \eg increasing network width, given a model architecture. Designing a new architecture can also lead to improved computational efficiency. However, the development cost of architecture design, via expert knowledge or neural architecture search \cite{NAS,NAS2}, makes direct comparison to scaling or ensembling difficult.
\subsubsection{Selective classification.}\label{sec:sc_res} 
\begin{figure}
    \centering
    \includegraphics[width=\linewidth]{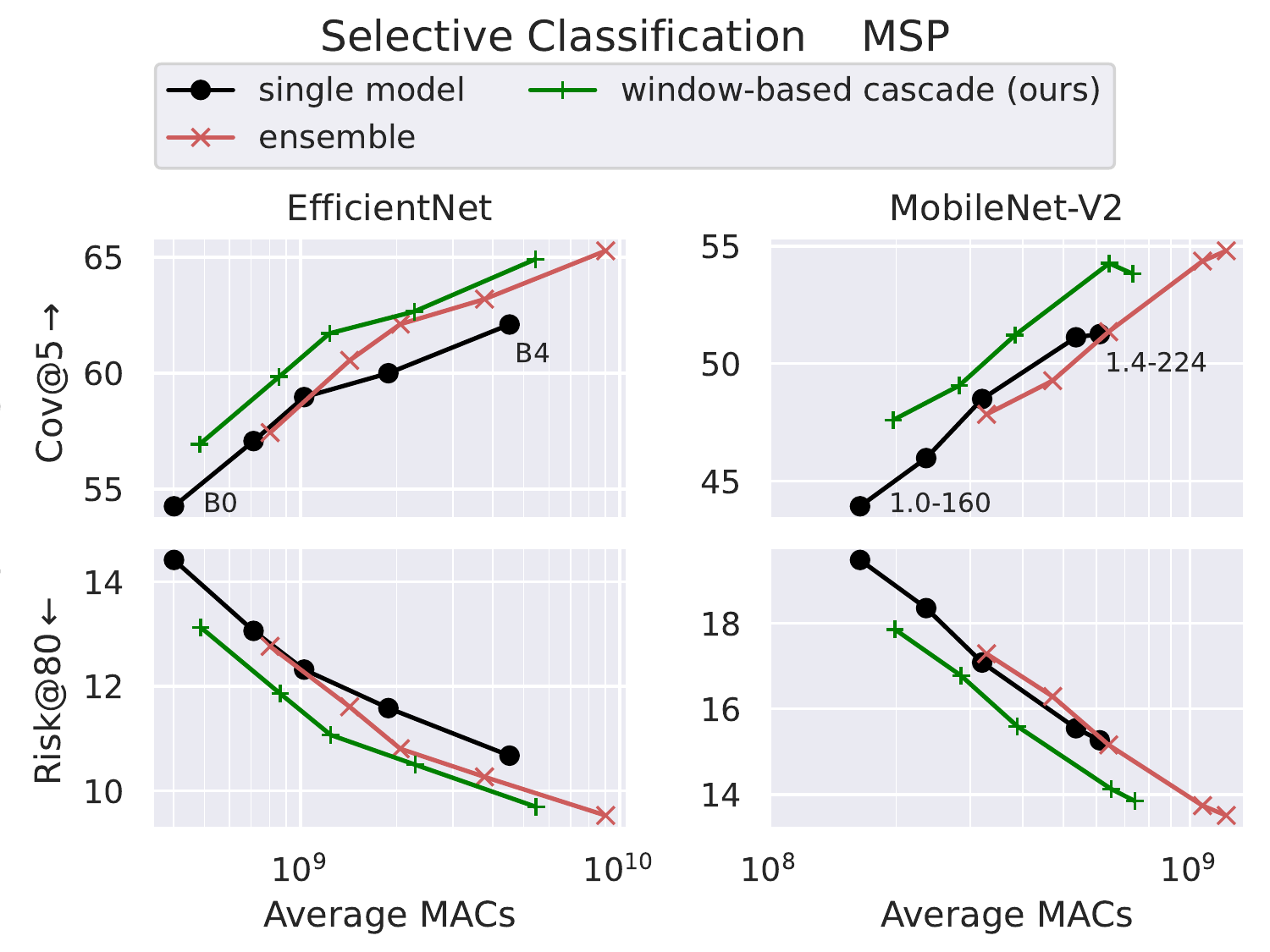}
    \caption{Selective classification performance against average computation (log-scale MACs) for EfficientNet B0$\rightarrow$B4 and scaled MobileNet-V2. Ensembles are quite competitive compared to single-model scaling, with an uncertainty-computation trade-off that is slightly worse for lower levels of compute and superior for higher computation. \textbf{Our \textcolor{ForestGreen}{window-based cascades} are able to achieve better uncertainty-computation trade-offs in all cases compared to scaling single models}. The data is ImageNet-1k.}
    \label{fig:sc}
\end{figure}

We evaluate SC performance at two different operating points on the risk-coverage curve: Cov@5$\uparrow$ and Risk@80$\downarrow$. These represent operating at a much lower risk tolerance vs without selection ($\sim$20-30\%) and operating at high coverage. Following \cref{sec:approach}, we find windows by expanding $[t_1, t_2]$ about $\tau$ of the first model in the cascade, capturing symmetric percentiles on either side.
We find on the validation set that passing $\pm10\%$ (20\% in total) of ID data around $\tau$ to the second stage is sufficient to recover most of the ensemble's performance. We use this window size for the remainder of our experiments. As mentioned previously, we leave further optimisation of $[t_1,t_2]$ to future work. 

We use the standard (negative) Maximum Softmax Probability (MSP) \cite{hendrycks17baseline,Geifman2017SelectiveCF} score, $U(\b x)= -\max_\omega P(\omega|\b x)$, to measure uncertainty for SC. For the ensemble, we use \cref{eq:ens_unc1} to calculate $U$. Note that this leads to better misclassification rejection compared to \cref{eq:ens_unc2}, as we are using \cref{eq:pred} for classification with the ensemble.
\paragraph{Cascades are best and ensembles are competitive.} 
\cref{fig:sc} shows that ensembles are slightly weaker than scaling single models in the lower computation region, and can outperform them at higher compute costs. This is in line with previous work on accuracy \cite{wang2022wisdom,Kondratyuk2020WhenES,powerlaw}. Moreover, it is clear, that by cascading ensembles using our window-based early-exit policy, a superior uncertainty-computation trade-off can be achieved, similar to accuracy in \cite{wang2022wisdom}. 
\subsubsection{OOD detection.}

We use two high-resolution OOD datasets designed for ImageNet-scale experiments. Openimage-O \cite{haoqi2022vim} and iNaturalist \cite{Huang2021MOSTS} are subsets of larger datasets \cite{openimages,inat} where images were selected to be disjoint to ImageNet-1k labels. The former covers a wide range of classes (like ImageNet); the latter contains botanical images. Additional dataset details can be found in \cref{app:data}. We evaluate at two operating points, FPR@80$\downarrow$ and FPR@95$\downarrow$, which represent high and very high retention of ID data. Initially, we set $[t_1,t_2]$ in the same fashion as for SC (\cref{sec:sc_res}). To measure average computation, we evaluate on a simple scenario where $\alpha=0.5$ (\cref{eq:mix}), i.e. the number of OOD samples is equal to ID samples. The OOD datasets are smaller, so we randomly subsample the ImageNet evaluation set.

We use the simple and popular Energy \cite{liu2020energy} score, $U(\b x)= -\log\sum_{k}\exp v_k$, where $\b v$ are the pre-softmax logits. 
For the ensemble, we use \cref{eq:ens_unc2} to calculate $U$ as in \cite{Xia2022OnTU}. 
\begin{figure}
    \centering
    \includegraphics[width=\linewidth]{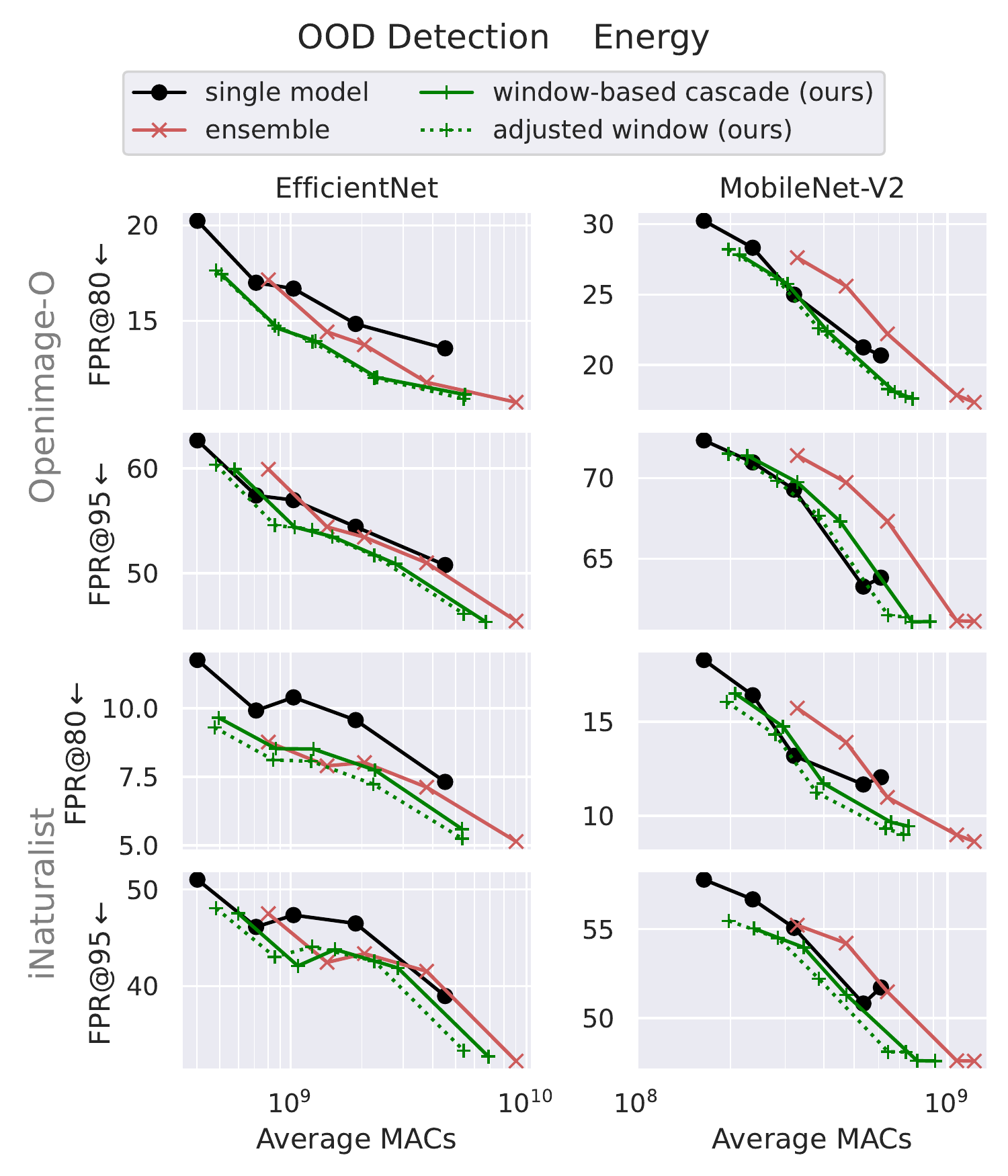}
    \caption{OOD detection performance vs computation. Window-based cascades are competitive overall but suffer from slowdown due to distributional shift in some cases. \textbf{Adjusting the window at deployment recovers cascade efficiency}, allowing for more efficient uncertainties compared to scaling single models. \textbf{Scaling single models does not reliably reduce FPR$\downarrow$, but ensembling does}, suggesting that ensembles/cascades are the \textit{safer} option for improving OOD detection. We assume a 1:1 ratio of ID:OOD data.}
    \label{fig:ood}
\end{figure}

\paragraph{Beware of distribution shift when early exiting.}
\cref{fig:ood} shows that in most scenarios, window-based cascades are better or competitive with single models. However, cascades are weaker in some cases. We hypothesise that this is due to there being more OOD data between $[t_1, t_2]$ compared to ID data, slowing the cascade down, as in this case $\tau$ is on the lower tail of the ID data. We remark that slowdown under distributional shift is a general problem with early-exit approaches, and is an interesting direction for further research. We also note that there will be less slowdown in scenarios where OOD data is rarer than ID data ($\alpha>0.5$). 

To correct for this slowdown, we propose \textit{adjusting} the window such that $[t_1,t_2]$ capture $\pm$10\% around $\tau$ of $p_{\text{mix}}$ rather than of $p_{\text{ID}}$ (\cref{eq:mix}). To achieve this, percentiles need to be calculated over the $U$s of the \textit{test} data. As this \textit{does not require labels}, it could be practically achieved by collecting statistics during deployment and adjusting $[t_1,t_2]$ accordingly. For online processing this may be done in a running manner, whereas for offline data processing, all samples could be processed by the first cascade stage, percentiles could be then calculated, and the corresponding subset of samples passed on to the next stage.  \cref{fig:ood} shows that adjusted windows recover cascade efficiency, allowing them to outperform or match single models in all cases. Larger speedups are observed for FPR@95$\downarrow$, as this $\tau$ is on the tail of the ID distribution of $U$ and is likely to be closer to the mode of the OOD distribution. 

\paragraph{Scaling doesn't reliably improve OOD detection but ensembling does.} \cref{fig:ood} shows that although model scaling has a general trend of reducing FPR$\downarrow$, it is inconsistent between different models, OOD datasets, and operating thresholds. In fact, \textit{scaling up does not always improve OOD detection}, \eg EfficientNet on iNaturalist. On the other hand, cascades/ensembles \textit{reliably} reduce FPR$\downarrow$ relative to their corresponding single members. This suggests that cascades are the \textit{safer} choice for efficiently boosting OOD detection performance, given that model scaling may not even improve detection at all.

Our results are roughly in line with those in \cite{vaze2022openset}, who find a general positive correlation between ID accuracy and OOD detection performance. We note that they observe more consistent improvements than us for OOD detection when model scaling with ResNets, however, the OOD data they use is different to ours.\footnote{It has been found in \cite{bitterwolf2023ninco} that the OOD datasets used in \cite{vaze2022openset} are heavily polluted with ID samples. However, it is unclear how this finding affects the corresponding empirical takeaways from \cite{vaze2022openset}.} Besides, they do observe a drop in performance when scaling from ResNet-50$\rightarrow$101 on their ``easy'' OOD dataset (\cite{vaze2022openset} Fig. 3b).
\subsubsection{SCOD.}
\begin{figure}
    \centering
    \includegraphics[width=\linewidth]{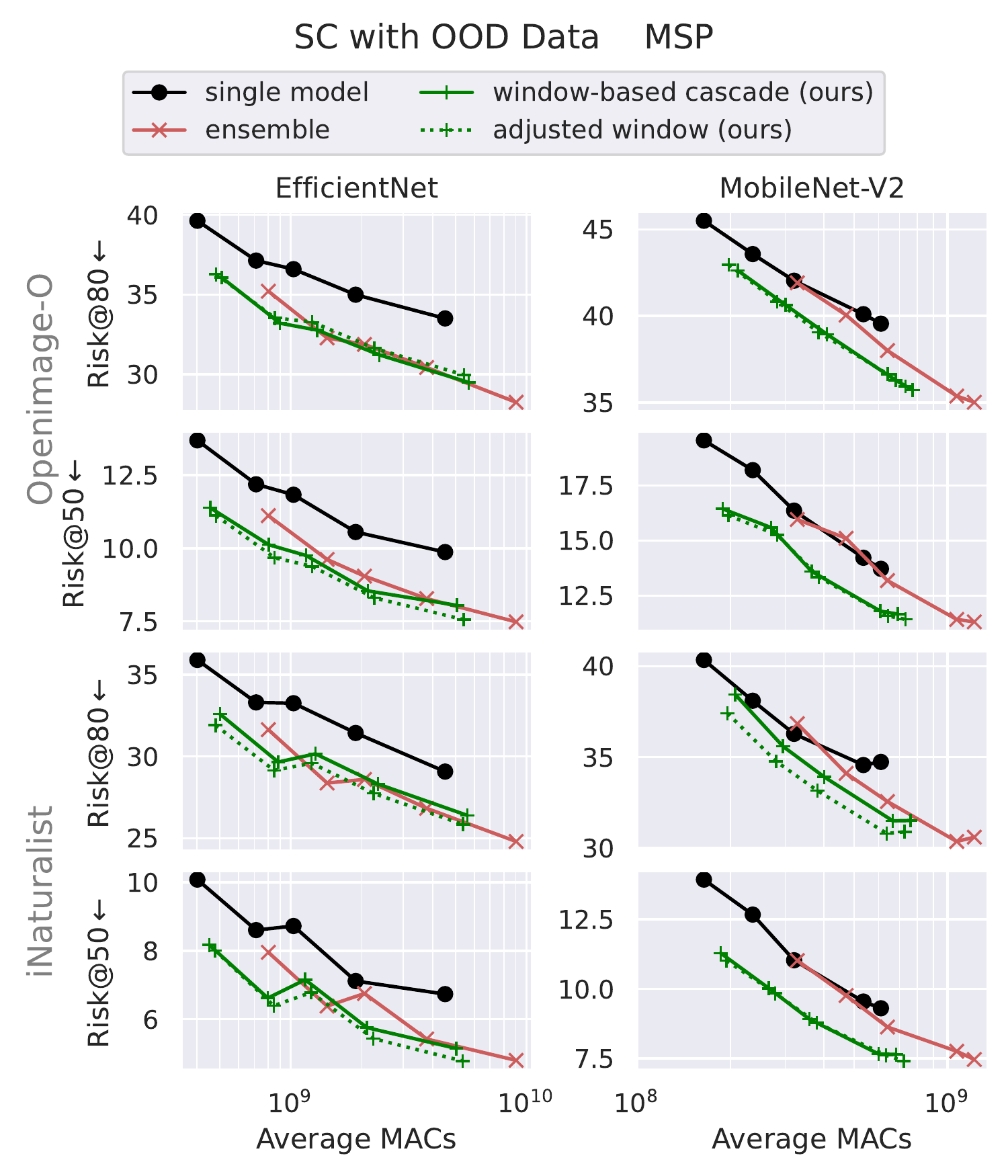}
    \caption{SCOD performance vs computation. Results are in-line with previous \cref{fig:sc,fig:ood}. \textbf{\textcolor{ForestGreen}{Window-based cascades} achieve more efficient uncertainty-related performance compared to scaling single models}, and sometimes benefit from deployment-time window adjustment. Ensembles perform better than before, being more efficient than single-model scaling for EfficientNets.} 
    \label{fig:scod}
\end{figure}
We keep the same OOD datasets as the previous section, as well as the 1:1 ratio of ID:OOD data. We set $\beta=1$ (\cref{eq:scodloss}), i.e. accepting OOD data and misclassifications is equally costly. We evaluate Risk@80$\downarrow$, and Risk@50$\downarrow$, where the coverage is over the ID data (see \cref{sec:tasks}). The first offers higher coverage, whilst the second results in much lower risks. We evaluate with windows set as in \cref{sec:sc_res}, as well as the adjusted windows from the previous section. We use (negative) MSP for $U$ just as in \cref{sec:sc_res}. 

Overall results (in \cref{fig:scod}) are similar to \cref{sec:sc_res}, with the cascades outperforming scaling single models for all levels of computation. Ensembles perform even better than for in-distribution SC, producing consistently better uncertainty-computation trade-offs vs scaling EfficientNet. This suggests that ensembling leads to even greater gains in being able to distinguish OOD data from \textit{correct} ID predictions, compared to rejecting ID misclassifications.

Some of the OOD detection behaviour from the previous section carries over to SCOD, with ensembling being a \textit{more reliable} way to reduce risk compared to model scaling (see EfficientNet on iNaturalist). Using adjusted windows leads to little change in Risk@50$\downarrow$, as a stricter $\tau$ will have fewer OOD samples near it to cause slowdown. Improvements in  Risk@80$\downarrow$, where $\tau$ is closer to the OOD distribution of $U$, can be observed in some cases, however.
\subsection{Real-world latency and throughput}\label{sec:real}
\begin{figure}
    \centering
    \includegraphics[width=\linewidth]{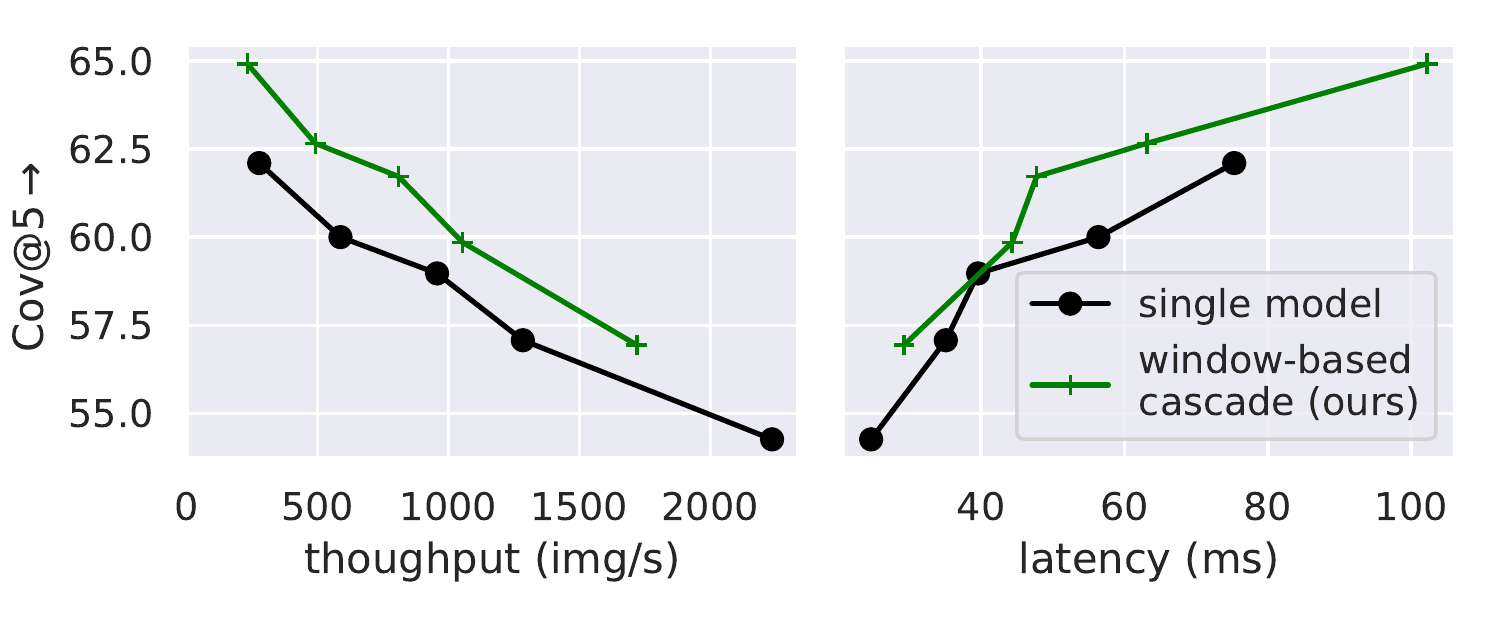}
    \caption{Real-world latency and throughput uncertainty efficiency gains for window-based cascades for EfficientNet-B0$\rightarrow$B4.  We use a NVIDIA V100 32GB GPU to measure throughput and an Intel Xeon E5-2698v4 CPU to measure latency.}
    \label{fig:rw}
\end{figure}
We validate our window-based cascades by measuring average real-world latency (batch size 1) and throughput (batch size 256) for EfficientNets. We use Pytorch and measure latency on an Intel Xeon E5-2698 v4 CPU and throughput on a NVIDIA V100 32GB GPU. For latency, the exit condition is evaluated for each sample, whilst for throughput, it is evaluated over the whole batch, and then the corresponding subset of the batch is passed to the second model. \cref{fig:rw} shows that the theoretical MACs results in \cref{sec:sc_res} can indeed translate to real-world deployment.

\subsection{Other early-exit architectures}
\begin{figure}
    \centering
    \includegraphics[width=\linewidth]{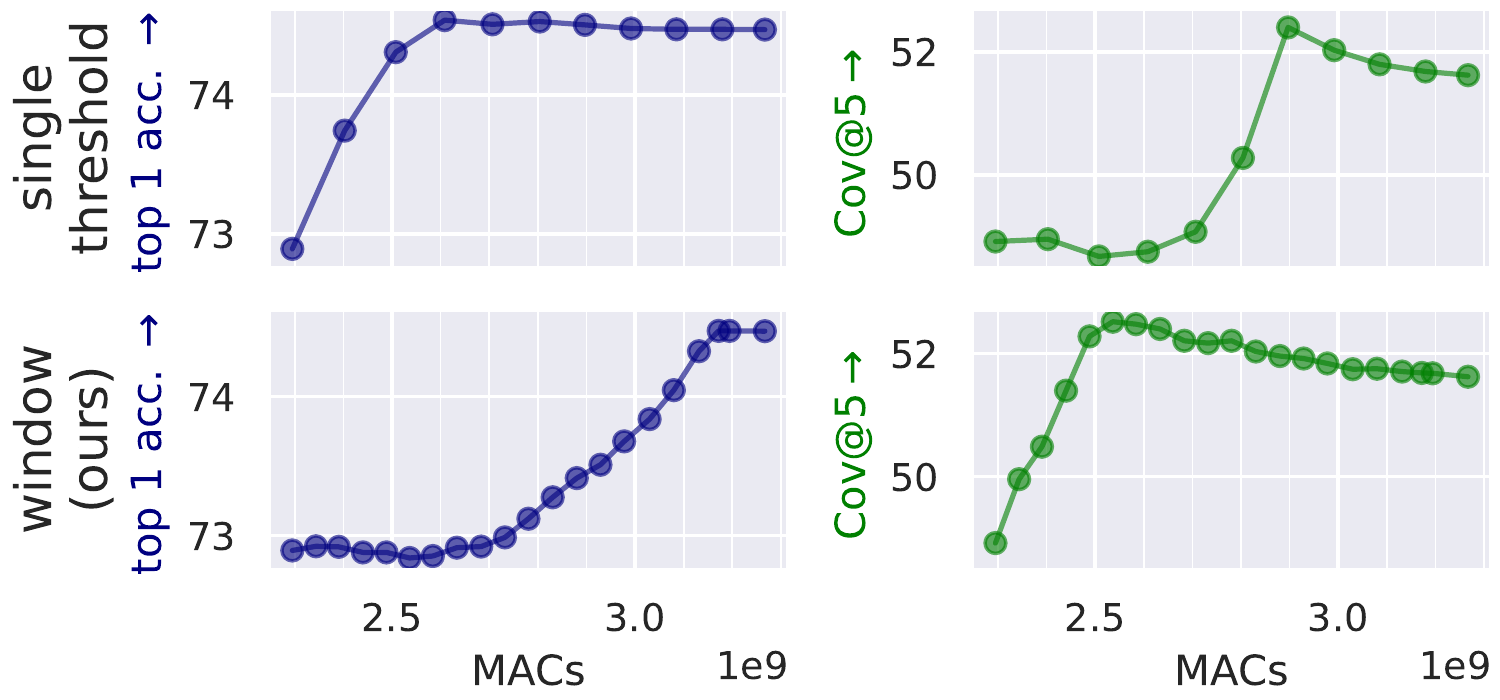}
    \includegraphics[width=.8\linewidth]{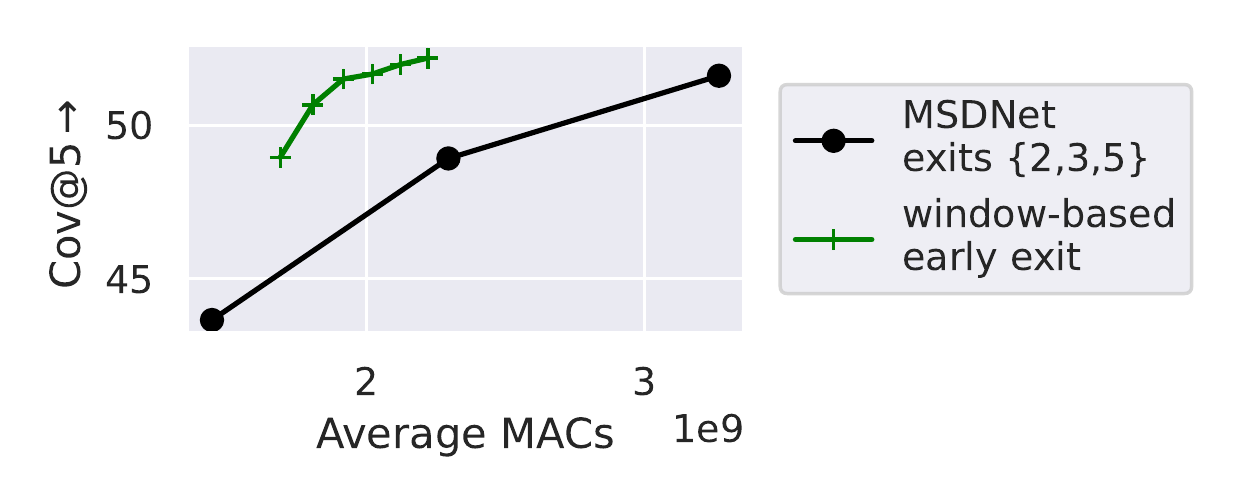}
    \caption{Top: comparison of single-threshold vs window-based early-exit policies on MSDNet exits 3$\rightarrow$5. \textbf{Our approach generalises to uncertainty estimation on other early-exit architectures.} Bottom: window-based policy applied to 3 exits of MSDNet for SC. \textbf{Our approach generalises to multiple early exits.}}
    \label{fig:msdnet_ee}
\end{figure}
Although the focus of this work has been on adaptive inference with cascaded Deep Ensembles, the proposed window-based early-exit policy can be applied to \textit{any} early-exit network. To show this, we apply a similar analysis to \cref{fig:strats} between the 3rd and 5th exits of a MSDNet\footnote{\url{https://github.com/kalviny/MSDNet-PyTorch}} (step=7) \cite{Huang2018MultiScaleDN} pretrained on ImageNet-1k. \cref{fig:msdnet_ee} shows that our window-based early-exit policy does indeed generalise to other early-exit architectures. Note that MSDNet exits are not ensembled \cite{Huang2018MultiScaleDN}. However, recall that SC can improve with either $g$ or $f$ (\cref{sec:tasks}), so we still expect later exits to be better as they are more accurate (better $f$).

To show that our approach generalises beyond a single exit, we also apply it to exits $\{2,3,5\}$ of MSDNet. We fix $[t_1,t_2]^{(2)}$ at $\pm10$ percentiles around $\tau^{(2)}$ and vary the width of the first window $[t_1,t_2]^{(1)}$. \cref{fig:msdnet_ee} shows that we achieve an efficient uncertainty trade-off vs the individual exits.  See \cref{app:ee} for similar results on additional early-exit architectures: GFNet \cite{gfnet} and ViT-based DVT \cite{rao2021dynamicvit}.

\subsection{Comparison with Packed Ensembles}

Packed Ensembles \cite{laurent2023packed} are a recent approach aimed at increasing the computational efficiency of Deep Ensembles. They utilise grouped convolutions in order to enforce explicit subnetworks within a larger neural network. This means that subnetworks are run in parallel, with the aim of better utilising the parallel compute found in modern machine learning hardware such as GPGPUs.

 We compare their SC performance on ImageNet to our approach in \cref{tab:packed}. Our window-based cascade is more efficient, both in terms of theoretical and practical computational cost. We note that Packed Ensembles are underwhelming in our experiments, which is likely due to there being less overparameterisation to exploit for ResNet-50 on ImageNet, compared to smaller-scale data such as CIFAR \cite{cifar}. On the other hand, our window-based cascades behave as expected, achieving near-Deep-Ensemble performance for $\sim 20\%$ additional cost (vs a single model).
\begin{table}[]
    \centering
   
\resizebox{\linewidth}{!}{
\begin{tabular}{@{}lllllll@{}}
\toprule
ResNet-50 ImageNet (-)MSP                             & \%cov & \%risk & MACs$\downarrow$  & GPU   & CPU  \\
Model Weights from \cite{laurent2023packed}                 &    @5$\uparrow$                  &   @80$\downarrow$                       &              ($\times10^9$)        &      Throughput$\uparrow$        & Latency$\downarrow$           \\
Uncertainty                       &                      &                          &        (av.)             &  batch=256                &   batch=1         \\ 
Pytorch Inference & & &   &(img/s)&(ms)\\
\midrule
Single Model                            & 54.2                 & 12.9                     & 4.09                & 826                    & 39.7                    \\
Deep Ens. $M=2$ (sequential)        & 57.5                 & 11.7                     & 8.18               & 439                  & 84.2                 \\
Window-Based Cascade  (ours)                 & 57.0                 & 11.9                     & 4.91               & 687                  & 50.6              \\
Packed Ens. $\alpha=3,M=4,\gamma=1$ & 53.7                 & 12.7                     & 9.29              & 335                & 67.6           \\ \bottomrule
\end{tabular}
}
\vspace{2mm}
     \caption{Comparison of our approach with the recently proposed Packed Ensembles \cite{laurent2023packed}. We use pre-trained ResNet-50 models provided by \citet{laurent2023packed}. \textbf{Window-based cascades achieve better selective classification performance at a lower theoretical and practical computational cost vs Packed Ensembles}. We perform measurements using the same hardware as in \cref{fig:rw}.}
    \label{tab:packed}
\end{table}

\section{Related Work}

\paragraph{Uncertainty estimation in deep learning.} There is a large body of work on improving predictive uncertainty estimates for deep learning approaches \cite{uncrev}. Some focus on improving performance on a specific binary detection task, such as SC \cite{Geifman2017SelectiveCF,ElYaniv2010OnTF}, OOD detection \cite{yang2021oodsurvey,hendrycks17baseline} and now most recently, SCOD \cite{Kim2021AUB, Xia_2022_ACCV}. Approaches include: alternative uncertainty scores \cite{liu2020energy, haoqi2022vim, Lee2018ASU, pmlr-v162-ming22a,vaze2022openset}, separate error detectors trained post-hoc \cite{SC_VQA, Kamath2020SelectiveQA}, modified training approaches \cite{du2022vos, Huang2021MOSTS, Hsu2020GeneralizedOD,Techapanurak_2020_ACCV}, feature clipping or sparsification \cite{Sun2021ReActOD, zhu2022boosting, sun2022dice} and confidence score combination \cite{haoqi2022vim, Xia_2022_ACCV, Huang2021OnTI,cen2023the}. Calibration \cite{pmlr-v70-guo17a} is a separate task where the aim is to produce probabilities that match the true accuracy of a classifier. It is orthogonal to the aforementioned tasks (and the focus of this paper), as in binary classification only the \textit{rank ordering} of scores matters, rather than absolute probabilities \cite{jaeger2023a}.

Other work aims to generally improve uncertainties over multiple evaluation metrics. Deep ensembles \cite{Lakshminarayanan2017SimpleAS} have stood out as a strong and reliable baseline in this field. However, due to their increased cost, there have been many proposed \textit{single-inference} approaches with the aim of improving uncertainty estimates at a lower cost. Examples include, low-rank weight matrices and weight sharing (BatchEnsemble) \cite{Wen2020BatchEnsemble:}, implicit/explicit subnetworks (MIMO/Packed Ensembles) \cite{havasi2021training,laurent2023packed}, using spectral normalisation combined with a gaussian process (SNGP) \cite{Liu2020SimpleAP} or a gaussian mixture model (DDU) \cite{Mukhoti2021DeterministicNN}, distribution distillation of the ensemble into a single model (EnD\textsuperscript{2}) \cite{EnDD,pmlr-v180-fathullah22a,logit-based-edd} and more \cite{LDU,pinto2022using,pinto2021mixmaxent,van2020uncertainty,Granese2021DOCTORAS}. We note that our work is generally orthogonal to these, as single-inference approaches can be ensembled \cite{Fort2019DeepEA} and then cascaded using our window-based adaptive inference approach. 
\paragraph{Early-exit networks.} Within the wide range of approaches for dynamic or adaptive inference \cite{dynamicsurvey}, where computation allocation is not fixed during deployment, early-exiting is a simple and popular choice \cite{Teerapittayanon2016BranchyNetFI,adaptive}. The core idea is to terminate computation early on ``easier'' samples in order to make more efficient use of computational resources. This can be performed by cascading entire networks of increasing predictive performance \cite{Wang2017IDKCF, wang2022wisdom,gfnet,rao2021dynamicvit}, or by attaching intermediate predictors to a backbone \cite{icml19shallowdeepnetworks,Kouris2021MultiExitSS} or by designing a specialised architecture \cite{Huang2018MultiScaleDN,Yang2020ResolutionAN}. Exit policies can be rules-based or learnt \cite{adaptive}. Although the vast majority of work in this field focuses on predictive accuracy, MOOD \cite{Lin_2021_CVPR} proposes an approach targeting OOD detection. Unlike our window-based policy, their policy relies on a heuristic based on input image PNG \cite{PNG} compression ratio to determine the exit and is restricted to only the task of OOD detection. \citet{pmlr-v158-qendro21a} utilise a multi-exit architecture to improve uncertainty-related performance. However, they ensemble all exits without any adaptive inference.
\section{Concluding Remarks}
In this work, we investigate the uncertainty-computation trade-off of Deep Ensembles vs scaling single models. Additionally, we propose a window-based early-exit policy and use it to efficiently cascade ensembles for uncertainty-related tasks such as selective classification, OOD detection, and SCOD. Evaluations on ImageNet-scale CNNs and data show that our window-based cascades achieve a superior uncertainty-computation trade-off compared to simply scaling up single models within an architecture family. We hope our work encourages more research into adaptive inference approaches for uncertainty estimation.

{\small
\bibliographystyle{ieee_fullname_natbib}
\bibliography{bibliography}
}
\clearpage
\appendix
\section{Additional Setup Details}\label{app:setup}
\paragraph{Training.}
We train two families of computationally efficient CNNs for image classification on ImageNet-1k \cite{Russakovsky2015ImageNetLS}: EfficientNet \cite{pmlr-v97-tan19a} and MobileNet-V2 \cite{Sandler2018MobileNetV2IR}. For EfficientNet, we scale width, depth and resolution as the original authors \cite{pmlr-v97-tan19a} from B0$\rightarrow$B4.  For MobileNet-V2 we use the scalings in \cref{tab:mob_scaling2}, which are taken from the Keras github repository.\footnote{\url{https://github.com/keras-team/keras/blob/master/keras/applications/mobilenet_v2.py}.} For each model we train a Deep Ensemble size $M=2$ using random seeds $\{1,2\}$ (everything is the same between ensemble members other than the random seed). This results in 10 individual ensembles, 5 composed of EfficientNets and 5 composed of MobileNet-V2s.
\begin{table}[h]

\small
\centering
\begin{tabular}{@{}llllll@{}}
\toprule
input resolution & 160 & 192 & 224 & 224 & 224 \\
width factor     & 1.0 & 1.0 & 1.0 & 1.3 & 1.4 \\ \bottomrule
\end{tabular}
\caption{MobileNet-V2 scaling used in experiments.}
\label{tab:mob_scaling2}
\end{table}

We train all models for 250 epochs on ImageNet-1k, and hold out a random subset of 50,000 images from the training set for validation (we then evaluate on the original validation set). We train using standard cross entropy. We use stochastic gradient descent with a learning rate of 0.2, weight decay of 4e-5 and a batch size of 1024 for all models other than EfficientNet-B4, for which we use a learning rate of 0.1 and batch size of 512 due to GPU memory constraints (following the scaling recommendations in \cite{Goyal2017AccurateLM}). We use cosine learning rate decay with a 5 epoch linear warmup.\footnote{This is a combination of the hyperparameters from \url{https://github.com/d-li14/mobilenetv2.pytorch} and the scaling approaches recommended in \cite{Goyal2017AccurateLM}.} We use default random resize-cropping and random horizontal flipping for data augmentation, and images are scaled using bicubic interpolation. 

 We train all of our models using PyTorch \cite{torch} and Lightning \cite{Falcon_PyTorch_Lightning_2019} distributed over 8 NVIDIA V100 32GB GPUs using Automatic Mixed Precision.\footnote{\url{https://developer.nvidia.com/automatic-mixed-precision}}
 Training and evaluation code can be found here: \url{https://github.com/Guoxoug/window-early-exit}. Please follow the instructions in the \texttt{README.md} file in order to reproduce our results and plots.
 \paragraph{Setting windows.}
 In general, we find $\tau$ on the validation set and then vary $[t_1, t_2]$ by placing them at increasing symmetric percentiles on either side of $\tau$. If either side of the window hits either the zeroth or 100th percentile, then the expansion will only apply to the other side, \eg if $\tau$ is set at TPR=95\% then there is only room for 5\% (of ID data) on the side more uncertain than $\tau$. As mentioned previously, we leave further optimisation of $[t_1,t_2]$ to future work. 
 \paragraph{Packed Ensembles.}
 For the experiments involving Packed Ensembles we use the same ResNet-50 models as those in Tab. 2 of \cite{laurent2023packed}, with weights kindly provided by \citet{laurent2023packed}. Implementation is the same as in the main experiments, and we treat Packed Ensemble outputs in the same way as (non-adaptive) Deep Ensembles.
\section{A Note on Uncertainty Scores $U$}
We remark that our approach is dependent on the compatibility of the uncertainty scores $U^{(1)}, U^{(2)},\dots ,U^{(M)}$ between different exits, as ultimately a single $\tau$ is used for the downstream uncertainty task. We find that in our experiments, simply using the same score method (e.g. Energy \cite{liu2020energy}) across all exits is sufficient. However, in a similar scenario, \citet{Lin_2021_CVPR} find it necessary to perform an additional score normalisation step. Although this may seem like common sense, we remark that we don't expect our approach to work if different exits use different score methods (e.g one exit uses MSP and another uses Energy), as these score methods may take very different absolute values.
\section{Additional Early-Exit Architectures}\label{app:ee}
\begin{figure}
    \centering
    \includegraphics[width=.8\linewidth]{msdnet_imagenet_cascade_earlyexit_cov5_macs_confidence.pdf}
    \includegraphics[width=.8\linewidth]{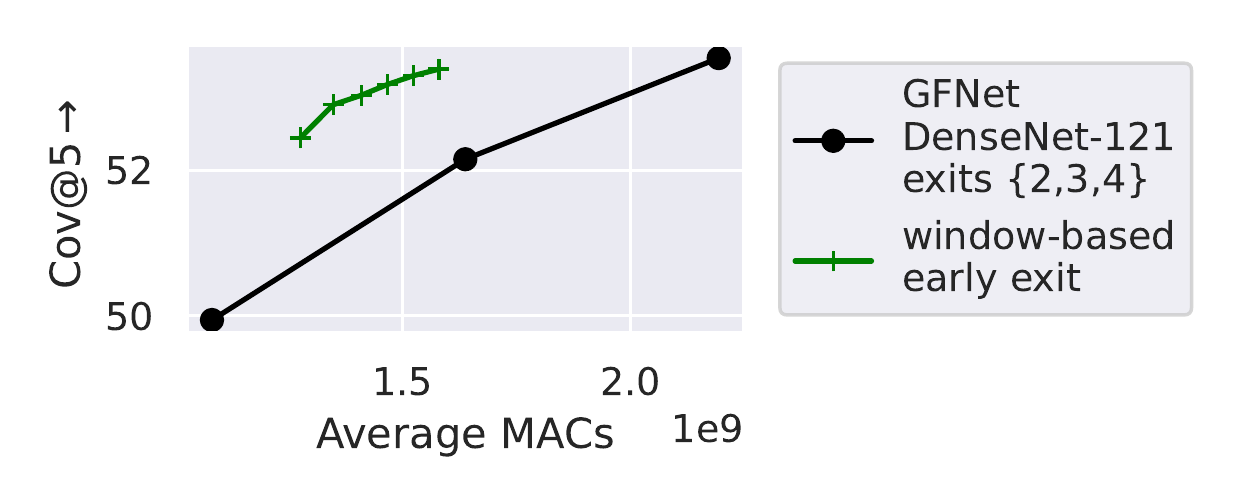}
    \includegraphics[width=.8\linewidth]{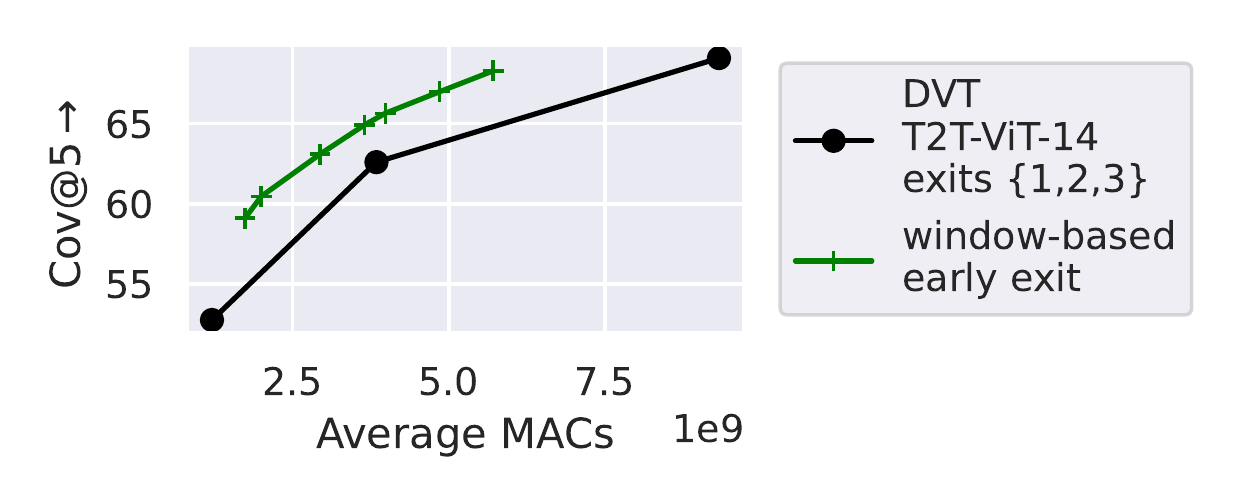}
    \caption{SC results for different early-exit architectures using 3 exits. Top: MSDNet, middle: GFNet, bottom: DVT. We achieve more efficient trade-offs compared to using individual exits.}
    \label{fig:add_ee}
\end{figure}
In addition to MSDNet \cite{Huang2018MultiScaleDN}, we also evaluate on GFNet \cite{gfnet} and the ViT-based DVT \cite{rao2021dynamicvit} in \cref{fig:add_ee}. These represent a diverse set of neural network architectures designed specifically for the early-exit paradigm. For all early-exit architectures we use publically available pretrained weights for ImageNet-1k.\footnote{\url{https://github.com/kalviny/MSDNet-PyTorch}\\ \url{https://github.com/blackfeather-wang/GFNet-Pytorch}\\ \url{https://github.com/blackfeather-wang/Dynamic-Vision-Transformer}} For MSDNet we use the version of the model designed for ImageNet with step=7, and experiment on a subset $\{2,3,5\}$ of the five available exits. For GFNet we use the model built on DenseNet-121 and exits $\{2,3,4\}$. For DVT we use the model built on T2T-ViT-14 and all 3 exits.
The percentiles used for $[t_1,t_2]$ are listed in \cref{tab:wind_width}:
\begin{table}[h]
    \centering
    \resizebox{\linewidth}{!}{
\begin{tabular}{@{}lll@{}}
\toprule
Early-Exit Arch. & \multicolumn{1}{c}{\%$\pm\tau^{(1)}$} for $[t_1,t_2]^{(1)}$ & \multicolumn{1}{c}{\%$\pm\tau^{(2)}$} for $[t_1,t_2]^{(2)}$\\ \midrule
MSDNet           & {[}10,15,20,25,30,35{]}               & {[}10,10,10,10,10,10{]}               \\
GFNet            & {[}10,15,20,25,30,35{]}               & {[}10,10,10,10,10,10{]}               \\
DVT              & {[}10,15,20,30,35,40,45{]}            & {[}0,0,10,10,10,15,20{]}              \\ \bottomrule
\end{tabular}
}
    \caption{Window widths for early-exit architectures.}
    \label{tab:wind_width}
\end{table}

For all three specialised early-exit architectures we achieve a better uncertainty-computation trade-off compared to using individual exits, validating our approach.

\section{Additional Exit Policy Comparisons}
\begin{figure}
    \centering
    \includegraphics[width=\linewidth]{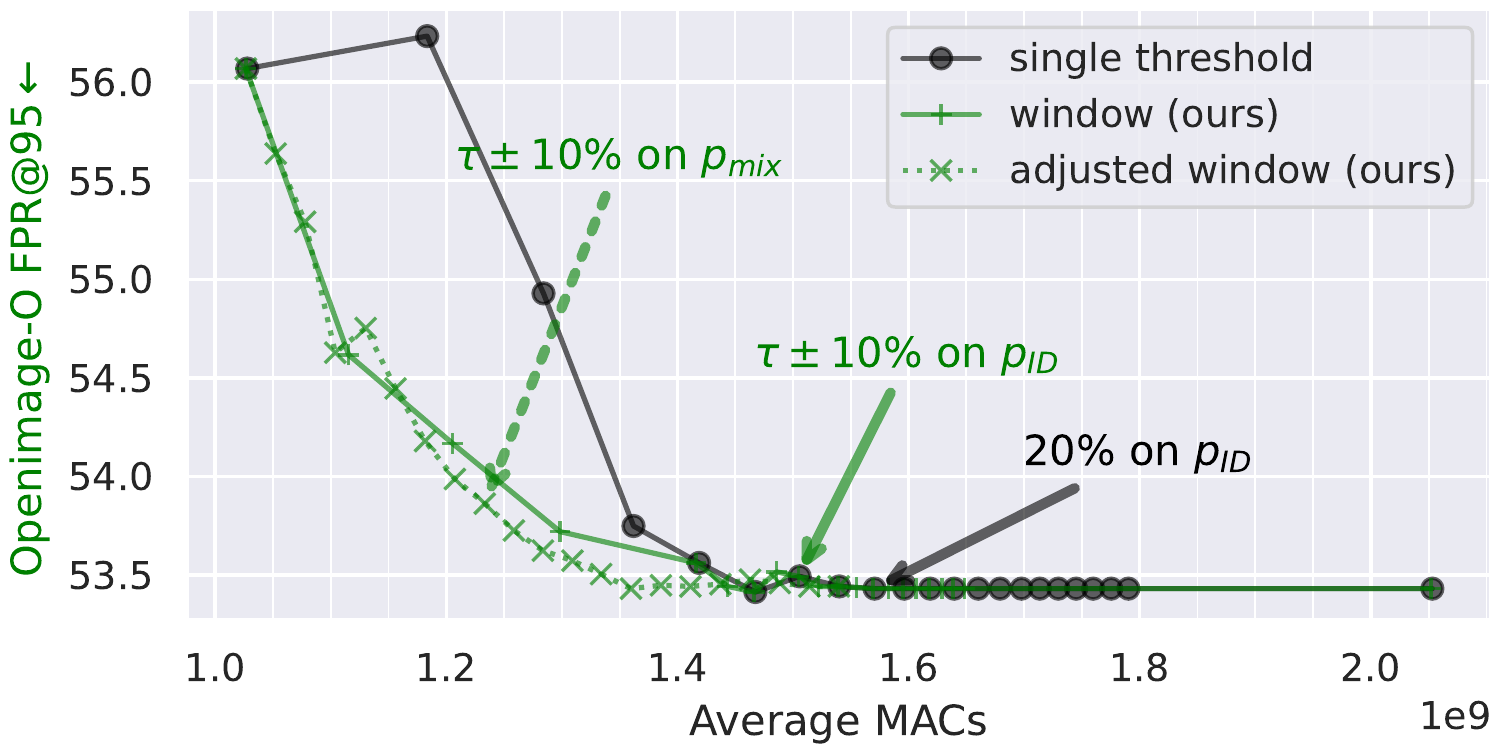}
    \caption{Comparison of OOD detection efficiency for different exit policies. The ratio of ID:OOD data is 1:1 ($\alpha=0.5$). Using a single threshold is inefficient compared to using a window. Adjusting $[t_1,t_2]$ using statistics from $p_{\text{mix}}$ significantly reduces the slowdown caused by distributional shift (if the window is set on $p_{\text{ID}}$). We use an EfficientNet-B2 ensemble with $M=2$.}
    \label{fig:ood_comp}
\end{figure}
We show comparisons between the single-threshold policy, and our window-based policy for OOD detection (FPR@95$\downarrow$, Openimage-O, $\alpha=0.5$). For the single-threshold and non-adjusted window approach we set $t$ and $[t_1,t_2]$ based on the percentage of $p_{\text{ID}}$ passed on to the next cascade stage. For the \textit{adjusted} window we set $[t_1,t_2]$ according to percentiles measured on $p_{\text{mix}}$ instead. 

Similarly to SC, \cref{fig:ood_comp} shows that the single-threshold approach only improves after $t$ (the exit threshold) passes over $\tau$ (the detection threshold). However, this happens earlier as the operating point TPR=95\% happens to be closer to the starting point of the single-threshold sweep (it starts from \textit{most} uncertain). Our window-based approach more efficiently improves OOD detection. 

Different specific exit policies are also marked. It can be seen that setting the window $[t_1,t_2]$ according to $p_{\text{mix}}$ rather than $p_{\text{ID}}$ significantly reduces slowdown caused by distribution shift. Setting $[t_1, t_2]$ to $\pm 10\%$ around $\tau$ on $p_{\text{ID}}$ leads to $\sim50\%$ of samples from $p_{\text{mix}}$ passing through to the second stage. Note that setting $[t_1,t_2]$ at $\pm 10$ percentiles around TPR=95\% on ID data only allows 15\% of ID data through as the window caps out on one side.

\section{Additional Selective Classification Results}
\begin{figure}
    \centering
    \includegraphics[width=\linewidth]{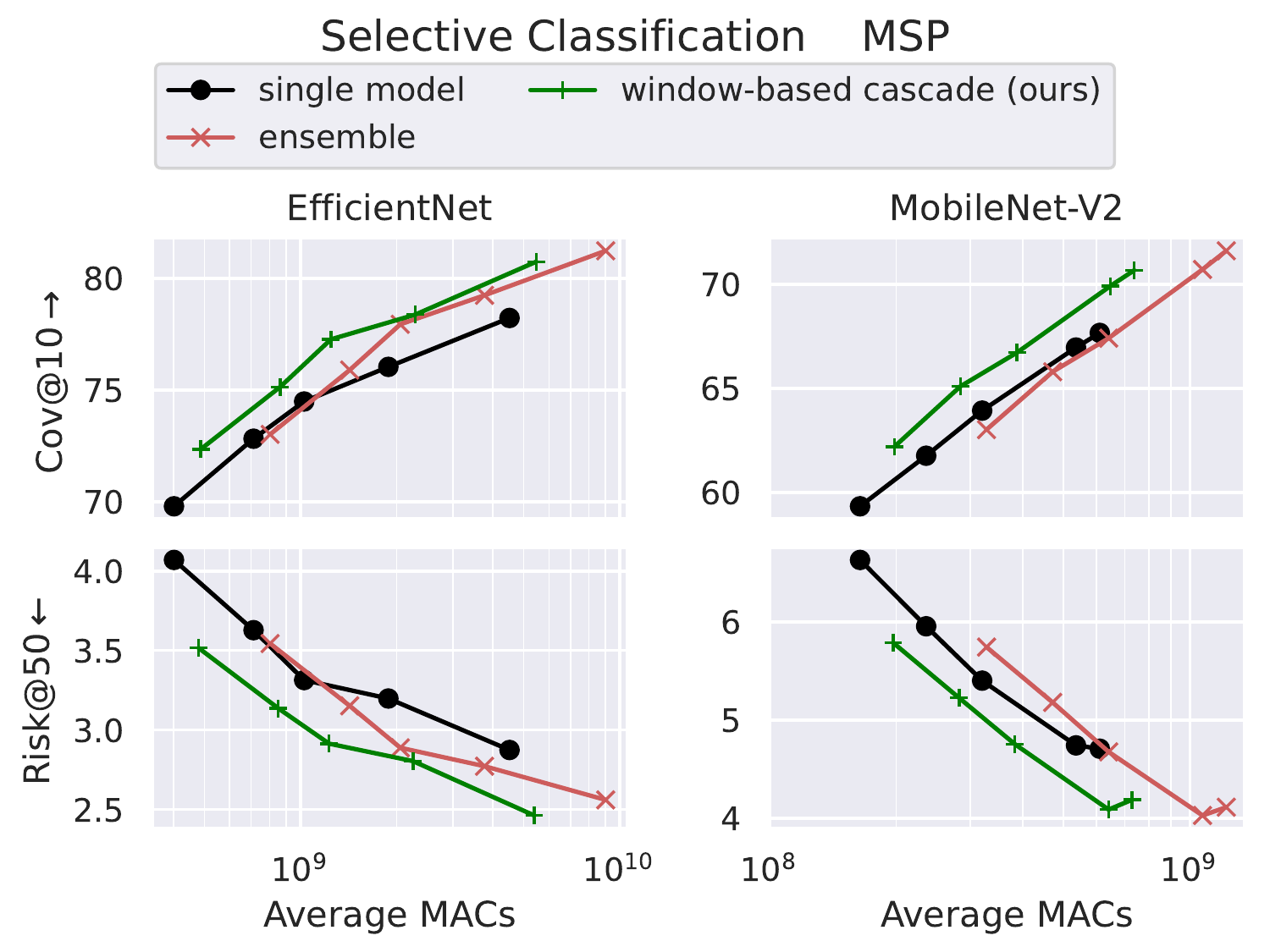}
    \caption{SC--computation comparison for single models, ensembles and window-based cascades, on additional operating thresholds. Results tell the same story as the main paper -- cascades are able to achieve the best uncertainty-computation trade-off.}
    \label{fig:sc_plus}
\end{figure}
We include additional SC results at two more operating points, Risk@50$\downarrow$ and Cov@10$\uparrow$ (\cref{fig:sc_plus}), which represent, compared to the main results, a lower coverage requirement and a higher risk tolerance respectively. The results are similar to those in the main paper, showing that cascades are able to achieve efficient uncertainty estimation compared to model scaling over a range of different operating thresholds.

\section{Accuracy-Computation Results}
\begin{figure}
    \centering
    \includegraphics[width=\linewidth]{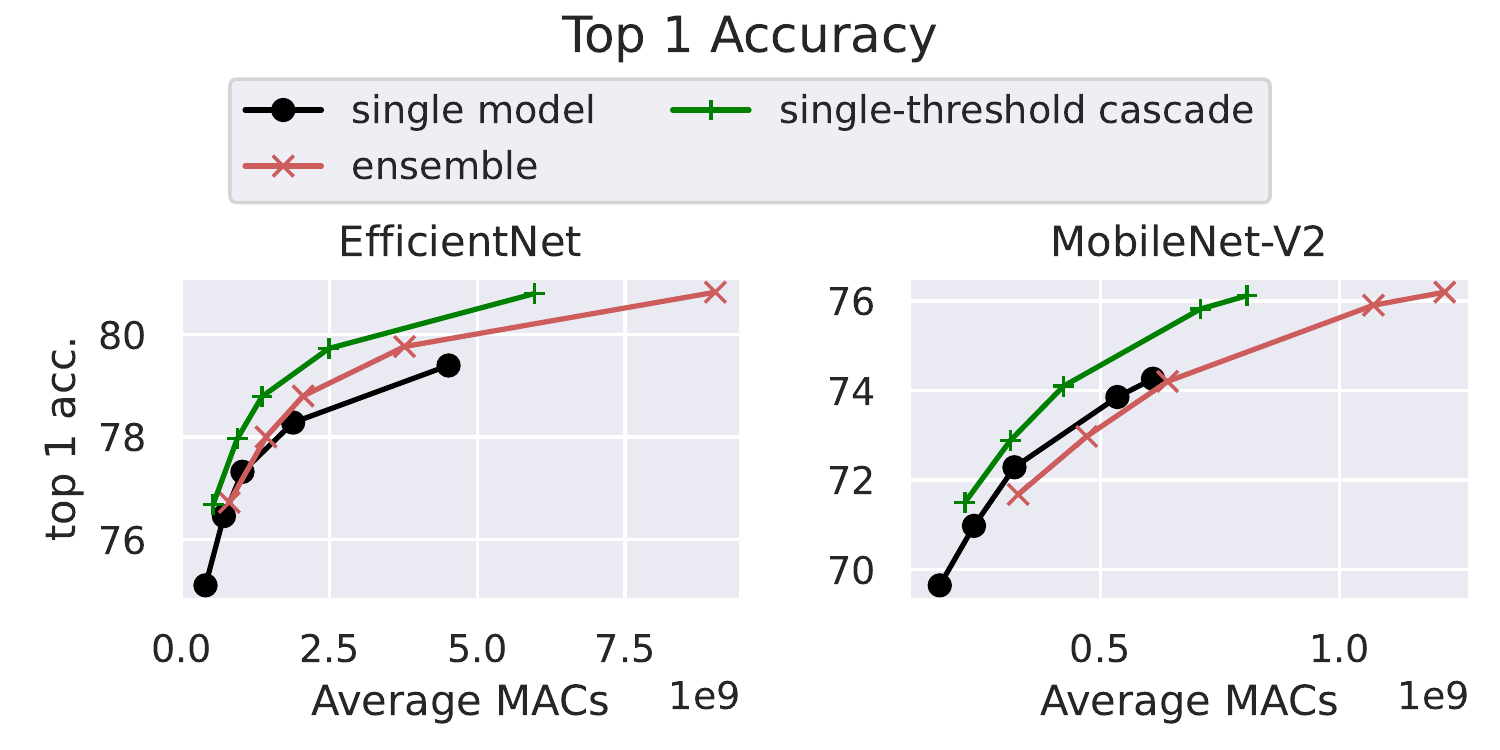}
    \caption{Accuracy--computation comparison for single models, ensembles and single-threshold cascades.}
    \label{fig:acc}
\end{figure}
\cref{fig:acc} shows the accuracy-computation trade-off using single-threshold cascades for EfficientNet and MobileNet-V2. We pass the most uncertain 20\% of samples from the 1st model to the second cascade stage. The results are unsurprisingly similar to those in \cite{wang2022wisdom}, i.e. ensembles are less efficient than single models in the low-compute region, but outperform them for higher computation levels. Cascades then allow ensembles to become more efficient for all levels of compute. We note the baseline accuracy of our EfficientNets is lower than in \cite{wang2022wisdom} as we train them for fewer epochs with a simpler recipe and larger validation set, however, we do not believe this affects the takeaways from our results. 
\section{Additional Dataset Information}\label{app:data}

\begin{figure*}[t]
    \centering
    \includegraphics[width=\textwidth]{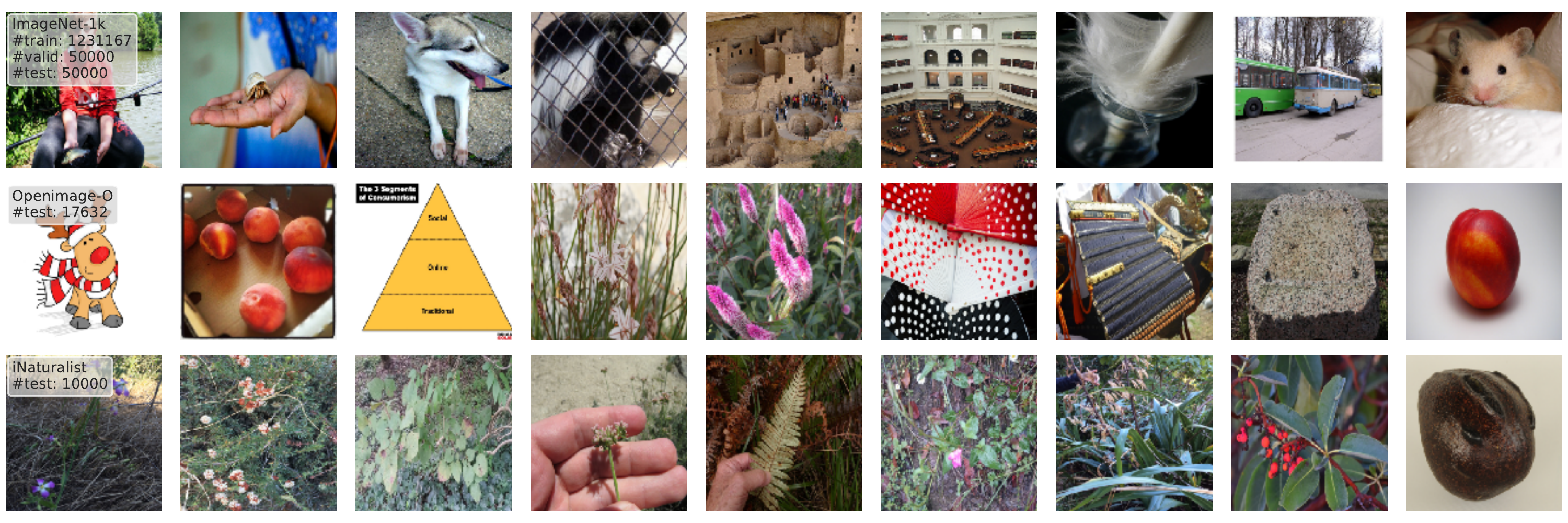}
    \caption{Example images from each image dataset used, with \#samples in each split.}
    \label{fig:data_examples}
\end{figure*}

We include randomly sampled images from the datasets used in this work: ImageNet-1k \cite{Russakovsky2015ImageNetLS}, Openimage-O \cite{haoqi2022vim,openimages}, and iNaturalist \cite{Huang2021MOSTS,inat}. We also show information about the number of samples in each dataset split (\cref{fig:data_examples}). The OOD datasets are recently released high-resolution benchmarking datasets. They aim to move vision-based OOD detection evaluation beyond CIFAR-scale images into more \textit{realistic} image-classification scenarios. The samples in each dataset have been carefully chosen to be semantically disjoint from the label space of ImageNet-1k. Openimage-O contains a wide range of classes like ImageNet, whilst iNaturalist contains botanical images. 
\section{Acknowledgements}
Guoxuan Xia is funded jointly by Arm ltd. and EPSRC. We would like to thank Alexandros Kouris, Pau de Jorge and Francesco Pinto for their helpful discussions and feedback. We would also like to thank Yulin Wang, Olivier Laurent and Gianni Franchi for kindly providing pre-trained weights for our experiments.

\end{document}